\documentclass[sigconf]{acmart}

\def\BibTeX{{\rm B\kern-.05em{\sc i\kern-.025em b}\kern-.08emT\kern-.1667em\lower.7ex\hbox{E}\kern-.125emX}}

\copyrightyear{2019} 
\acmYear{2019} 
\acmConference[SC '19]{The International Conference for High Performance Computing, Networking, Storage, and Analysis}{November 17--22, 2019}{Denver, CO, USA}
\acmBooktitle{The International Conference for High Performance Computing, Networking, Storage, and Analysis (SC '19), November 17--22, 2019, Denver, CO, USA}
\acmPrice{15.00}
\acmDOI{10.1145/3295500.3356147}
\acmISBN{978-1-4503-6229-0/19/11}

\usepackage{scalerel}
\newcommand\vfrac[2]{\ThisStyle{%
  \setbox0=\hbox{$\SavedStyle#1#2$}%
  \setbox2=\hbox{$\SavedStyle X$}%
  \ifdim\ht0>\ht2\setlength{\ht0}{\ht2}\fi%
  #1\mathord{\stretchto{\raisebox{2.3\LMpt}{$\SavedStyle/$}}{\ht0}}#2}}
\usepackage{url}
\usepackage{mathrsfs}
\usepackage{booktabs} 
\usepackage{setspace}
\usepackage{subcaption}
\usepackage{verbatim}

\usepackage{courier}
\usepackage{listings}
\usepackage{enumitem}
\usepackage{fixltx2e}
\usepackage[colorinlistoftodos]{todonotes}

\usepackage[most]{tcolorbox}
\usepackage{todonotes}
\usepackage{xcolor}
\usepackage{color}

\usepackage{xr}
\definecolor{mygreen}{rgb}{0,0.6,0}
\usepackage{multirow}
\usepackage{float}
\usepackage{graphicx}
\usepackage{caption}
\usepackage{algorithm}
\usepackage{algorithmic,tabularx}

\usepackage[utf8]{inputenc}

\usepackage{cleveref}
\crefname{section}{§}{§§}
\Crefname{section}{§}{§§}
\usepackage{amsmath,bm}

\setcopyright{acmlicensed}

\usepackage{caption}
\newlength\myindent
\newcommand\bindent{%
  \begingroup
  \setlength{\itemindent}{\myindent}
  \addtolength{\algorithmicindent}{\myindent}}
\newcommand\eindent{\endgroup}

\newlength\wqindent
\newcommand\bwqindent{%
  \begingroup
  \setlength{\itemindent}{\wqindent}
  \addtolength{\algorithmicindent}{\wqindent}}
\newcommand\ewqindent{\endgroup}

\makeatletter
\newcommand{\multiline}[1]{%
  \begin{tabularx}{0.9\dimexpr\linewidth-\ALG@thistlm}[t]{@{}X@{}}
    #1
  \end{tabularx}
}
\makeatother

\begin{document}

\author{Wenqian Dong}
\affiliation{
  \institution{University of California, Merced}
}
\email{wdong5@ucmerced.edu}

\author{Jie Liu}
\affiliation{
  \institution{University of California, Merced}
}
\email{jliu279@ucmerced.edu}

\author{Zhen Xie}
\affiliation{
  \institution{University of California, Merced}
 }
\email{zxie10@ucmerced.edu}

\author{Dong Li}
\affiliation{
  \institution{University of California, Merced}
 }
 \email{dli35@ucmerced.edu}
 
\title{Adaptive Neural Network-Based Approximation to  Accelerate Eulerian Fluid Simulation}

\renewcommand{\shortauthors}{W . Dong et al.}

%
\begin{abstract}

The Eulerian fluid simulation is an important HPC application. The neural network has been applied to accelerate it. The current methods that accelerate the fluid simulation with neural networks lack flexibility and generalization. In this paper, we tackle the above limitation and aim to enhance the applicability of neural networks in the Eulerian fluid simulation. We introduce Smart-fluidnet, a framework that automates model generation and application. Given an existing neural network as input, Smart-fluidnet generates multiple neural networks before the simulation to meet the execution time and simulation quality requirement. During the simulation, Smart-fluidnet dynamically switches the neural networks to make best efforts to reach the user’s requirement on simulation quality. Evaluating with 20,480 input problems, we show that Smart-fluidnet achieves 1.46x and 590x speedup comparing with a state-of-the-art neural network model and the original fluid simulation respectively on an NVIDIA Titan X Pascal GPU, while providing better simulation quality than the state-of-the-art model.
\end{abstract}



\maketitle

\section{Introduction} 
\label{sec:intro}

The fluid simulation aims to study the flow of fluid materials and has been widely applied to multiple disciplines such as chemical physics and material science~\cite{de2012computational,choi2018feasible,rutkowski2018surgical}. However, the simulation of fluid dynamics usually requires prohibitively high computational resources~\cite{xia2019gpu,snyder2019cfd} and thus limits its application in the related fields. 

The neural network-based machine learning model, as a tool to learn and model complicated (non)linear relationships between input and output datasets, has shown preliminary success in HPC problems (e.g., detecting neutrinos~\cite{ichep16:radovic}, developing Bose-Einstein Condensates state~\cite{sr15:widley}, and recognizing extreme weather events~\cite{nips17:racha}). Using neural networks, scientists are able to augment existing simulations by improving accuracy and significantly reducing latency. 

The neural network has been applied to accelerate fluid simulation as well~\cite{kim2019deep,tompson2017accelerating,yang2016data}.
By replacing some execution phases with neural networks, the most recent work reports $14.6 \times$ to $716 \times$ performance improvement~\cite{tompson2017accelerating}.

However, the current methods to accelerate the fluid simulation using neural networks have fundamental limitations. 
First, the current method to apply the neural work to the fluid simulation lacks flexibility. In particular, given the simulation code, the current method usually generates just one neural network model. At runtime, this model is used throughout the whole execution to approximate some target computation.  This method ignores the fact that replacing the target computation at different execution phases of the fluid simulation can have different implications on the simulation quality. At some execution phase, using another neural network model may be able to generate higher simulation quality without losing performance. Hence, using multiple neural network models instead of one is a better strategy. However, with just one single neural network model, the current method lacks such flexibility.

Second, the current method to apply the neural network to the fluid simulation lacks a generalization ability. In particular, given a fluid simulation code with the neural network applied, there is no guarantee that the application can run the simulation to completion and meet the requirement on simulation quality for every input problem. Using a single neural network model for all input problems either leaves the performance opportunities on the table (discussed in the last paragraph), have large simulation quality violations, or both.

Third, there is no \textit{systematic} approach to construct and apply neural network models to the fluid simulation. How to construct neural networks to meet the user requirement on performance (execution time) and simulation quality is challenging. Currently, domain scientists build neural networks intuitively. There is no systematic approach to help them build and apply neural networks for HPC applications. The recent work on Auto-Keras~\cite{jin2018efficient} and AutoML~\cite{escalante2019automl} aims to automatically generate a neural network model with high accuracy. However, they lack concerns on high performance (execution time), and focus on image processing or natural language processing. Hence, they are not directly usable by HPC.

In this paper, we tackle the above limitations and aim to enhance the applicability of neural networks in HPC applications (particularly the Eulerian fluid simulation). 
Given an existing neural network model as input, our system uses a systematic approach to construct multiple neural network models and dynamically switches them at runtime during the execution of the fluid simulation to meet the user requirements on performance and simulation quality. 

In order to tackle the above limitations, we must address three challenges. First, we must automatically generate multiple neural network models to enable high flexibility and better generality when applying neural networks.
Given the user requirements on performance and simulation quality, we aim to generate multiple neural network models, each of which has different topologies and different implications on performance and simulation quality. We should not expect the domain scientists to manually construct models.

Second, how to select neural network models at runtime to enable the best performance without violating the quality requirement is a challenge. We must have a method to predict the impact of applying a neural network model at a certain execution phase on the final simulation quality. 

Third, a neural network model can approximate the fluid simulation with high accuracy for some input problems but not for all. 
How to construct neural network models to provide a high-quality approximation for a large number of input problems and ensure overall performance benefit is another challenge. 

In this paper, we focus on a Eulerian fluid dynamic simulation code (mantaflow~\cite{thuerey2016mantaflow}) and introduce a framework (named ``Smart-fluidnet'') to address the above three challenges. Smart-fluidnet has three major components. (1) It includes a model construction plugin for Auto-Keras to extend its functionality to enable automatic construction of multiple neural network models. (2) Smart-fluidnet also includes a multilayer perceptrons model (MLP) to guide the model selection process to meet the requirement on performance and simulation quality. The neural network models selected by MLP is able to cover more input problems to ensure overall performance benefit. (3) Smart-fluidnet includes a runtime component integrated into mantaflow and dynamically switches the neural network models to make best efforts to meet the user requirement on simulation quality. The runtime component is based on a metric and a lightweight runtime algorithm that can predict the final simulation quality in the middle of the fluid simulation. 

We summarize the major contributions of this paper as follows:
\begin{itemize}

\item A systematic approach and a framework (Smart-fluidnet) to accelerate the Eulerian fluid simulation; Our evaluation shows that using 20,480 input problems for the simulation, Smart-fluidnet achieves 46\% performance improvement over the Tompson's model~\cite{tompson2017accelerating} (a state-of-the-art neural network model) on average and $590\times$ speedup over the original fluid simulation on average, while providing better simulation quality than the Tompson's model.
\item A new methodology that constructs, selects, and applies multiple neural network models (instead of one) to address the fundamental limitation of model flexibility and generalization in the existing neural network-based approximation. We demonstrate great potential of using this methodology to meet user requirements on the simulation quality and execution time.
\end{itemize}

\section{Background} 
\label{sec:challenge}
In this section, we provide background on the Eulerian fluid simulation and neural network-based approximation. 
{In the rest of the paper, the term \textit{performance} means execution time, not prediction accuracy in the machine learning field. We also use the terms \textit{approximation model} and \textit{neural network model} interchangeably.}

\subsection{Eulerian Fluid Simulation}
The Eulerian fluid simulation, in essence, solves the Navier-Stokes equations. The Navier-Stokes equations describe the fluid movement under a continuous velocity field $\vec{u}$ and a pressure field $p$. When the fluid has zero viscosity, one uses the incompressible Navier-Stokes equations, which can be expressed as the Euler equations as follows:

\begin{equation}
   \frac{\partial \vec{u}}{\partial t} = - \vec{u}  \cdot \bigtriangledown \vec{u} - \frac{1}{\rho} \bigtriangledown p    + \vec{g},
   \label{equ:NS_momentum}
\end{equation}
\begin{equation}
       \bigtriangledown \cdot \vec{u} = 0
       \label{Equ:incompressibility_condition}
\end{equation}
Equation \ref{equ:NS_momentum} is a vector equation called ``momentum equation''. This equation can make the velocity field stay divergence-free. Equation \ref{Equ:incompressibility_condition} is the incompressibility condition, which enforces fluid volume to remain constant throughout the simulation.
In the above two equations, $t$ is time, $\rho$ represents fluid density, and $\vec{g}$ represents gravity.

Mantaflow performs fluid simulation by iteratively solving Equations~\ref{equ:NS_momentum} and~\ref{Equ:incompressibility_condition}. 
In this paper, we use MAC (marker-and-cell) grids~\cite{harlow1965numerical} to discretize fluid flows, and use the finite difference (FD) method to calculate partial derivative on each grid~\cite{jeffrey2003applied,yu2005dlm}.

For each velocity component that borders a grid cell, the FD method iteratively applies updates on velocity and pressure.  In a grid cell, the pressure is sampled at the grid cell center and the velocity is sampled at the centers of the vertical faces of the grid cell. The above method is common and can simplify the handling of solid-cell boundaries conditions.

To solve Equations~\ref{equ:NS_momentum} and~\ref{Equ:incompressibility_condition}, mantaflow uses a standard operator splitting method~\cite{steger1981flux,yanenko1971simple,chorin1968numerical} to split up Equation~\ref{equ:NS_momentum} (Equation~\ref{Equ:incompressibility_condition} is used as a constraint) into three parts. 
The three parts are advection, adding external force, and pressure projection.
Algorithm~\ref{Eulerian algorithm} depicts the implementation of the Eulerian simulation 
in mantaflow, which includes the above three parts.

\begin{algorithm}
\caption{Velocity Update in the Euler Equation}
\label{Eulerian algorithm}
\begin{algorithmic}[1]
\REQUIRE Simulation time step $N$; 
\STATE Start with an initial divergence-free velocity field $\vec{u}^0$\\
\STATE Determine a good time step ${\bigtriangleup t}$ to go from time $t_n$ to time $t_{n+1}$.\\
\FOR{ $n \gets 1$ to $N$ }        
\STATE \textbf{Advection.} Set $\vec{u}^{A}$ = advect($\vec{u}^{n}$, $\Delta t$, $q$);\\
\STATE \textbf{Add body force.} $\vec{u}^B = \vec{u}^A +\bigtriangleup t \vec{f}$; \\
\STATE \textbf{Pressure projection.} set $\vec{u}^{n+1}$ = Project(${\bigtriangleup t}$, $\vec{u}^{B}$):
\bindent
\STATE  1) Solve the Poisson eq. $\bigtriangledown \cdot \bigtriangledown \vec{p_n} = \frac{1}{\bigtriangleup t} \bigtriangledown \cdot \vec{u}^B$ 
\bwqindent
\STATE //Use a PCG solver to to update $\vec{p}_n$. \\
\STATE  Set initial guess $\vec{p}_n$=0 and residual vector $r$=$d$ (if $r$=0, then return $\vec{p}_n$)\\
\STATE  Set search direction $\vec{s} = ApplyPreconditioner(r) $;
 \STATE \textbf{while} residual doesn't reach the convergence criteria
 \STATE  \textbf{do}

    \STATE  \hskip1em  Set $\alpha = \frac{r^{T} r}{s^{T}A s}  $;
   \STATE  \hskip1em  Calculate the residual $r = r - \alpha A \vec{p}_n$;
   \STATE  \hskip1em  Update the solution $\vec{p}_n = \vec{p}_n + \alpha \vec{s}$;
    \STATE \hskip1em  Update the conjugated direction $\vec{s} = r + \beta \vec{s}$; 

\STATE \textbf{end while}
\ewqindent
\STATE  2) Apply velocity update: $\vec{u}^{n+1} = \vec{u}^{B} - \Delta t \frac{1}{\rho} \bigtriangledown \vec{p}_n$;
 \eindent
\ENDFOR
\RETURN{0} 
\end{algorithmic}
\end{algorithm}

The Eulerian simulation in mantaflow includes $N$ time steps.
The first part (Lines 4-5) of each time step is to solve the momentum equation to get an auxiliary velocity field $\vec{u}^B$. $\vec{u}^B$ is a velocity approximation which is not divergence-free, and the pressure-gradient term ($\bigtriangledown p$) used during the solving process is computed in the previous time step. At Line 7, the divergence-free pressure $\vec{p_n}$ is computed by solving a Poisson's equation which includes the divergence of $\vec{u}^B$ and a scaled gradient of the pressure. At Line 18, a divergence-free velocity field, $\vec{u}^{n+1}$ is calculated, by subtracting off the pressure gradient from the approximate velocity field $\vec{u_B}$. In the above process, solving the Poisson's equation is the most crucial and time-consuming step to preserve the divergence-free constraint on the velocity and maintain simulation accuracy.

The process of solving the Poisson's equation in mantaflow (Line 7 in Algorithm~\ref{Eulerian algorithm}) is based on the Preconditioned Conjugate Gradient (PCG) method, which involves large computation that iteratively converges to  meet a convergence criteria (Lines 8-17).
Mantaflow uses a multi-grid approach~\cite{mcadams2010parallel} as a preprocessing step of the PCG method. The pre-conditioner (Line 10) applied in mantaflow is the Modified Incomplete Cholesky L0 preconditioner, called ``MICCG(0)''. In this paper and the existing work~\cite{tompson2017accelerating}, neural networks are used to approximate this PCG method.

In this paper, we simulate a 2D smoke plume~\cite{genevaux2003simulating,fedkiw2001visual}. The simulation output in mantaflow is a smoke \textit{dense matrix} of a rendered smoke frame. The smoke dense matrix represents density blurring of the plume, which reflects fluid movement. After using neural networks to approximate the computation in the fluid simulation, the output dense matrix can be different from that in the original simulation (using the mantaflow's PCG-based solver), which means we have quality loss. The simulation quality loss ($Q_{loss}$) is formally defined by: 
\vspace{-10pt}
\begin{equation}
\label{eq:quality_loss}
    Q_{loss} =\frac{1}{N\times M}\sum_{i=1}^N\sum_{j=1}^M {\left|\rho_{ij}^{*} - \rho_{ij} \right|} ,
\end{equation}
where $\rho$ refers to the smoke density matrix generated in the original simulation, and $\rho^{*}$ represents the smoke density matrix generated after applying neural network-based approximation. $\rho_{ij}$ and  $\rho^{*}_{ij}$ are matrix elements with the coordinate ($i, j$). In essence, Equation~\ref{eq:quality_loss} calculates the average relative error of all matrix elements. After applying the neural network-based approximation, we want to avoid quality loss (i.e., we do not want to lose simulation accuracy). 

\subsection{Neural Network-Based Approximation}
The neural network is a general-purpose method that can be used to learn and model complicated linear and non-linear relationships between input and output datasets. 
Hence, the neural network has been used to approximate some conventional algorithms in an application to improve performance~\cite{kim2019deep,tompson2017accelerating,yang2016data}. 
The neural networks are expected to generate similar outputs as the conventional algorithms when fed with the same inputs as for the conventional algorithms. 

A neural network can be represented as a directed acyclic graph where nodes of the graph are connected neurons. Embedded in the graph, there are a number of parameters (or ``weights''). Those neurons and weights are organized as layers. 
The process of obtaining the values of those weights are called  \textsl{training}. Once the neural network is trained offline using training datasets, it can be used online within the fluid dynamic simulation to improve performance. 
During training, an \textsl{objective function} is used to calculate the model accuracy loss, so that the weights can be adjusted to reach better model accuracy.

In this paper, we use Convolutional Neural Networks (CNN) to approximate computation (i.e., using the PCG solver to solve the Poisson's equation) in the Eulerian fluid simulation. This computation is the most time-consuming part of the Eulerian fluid simulation. Our profiling results reveal that this computation takes 70-80\% of total simulation time. 

Recent work~\cite{tompson2017accelerating} introduces an unsupervised learning framework to generate a CNN model with five stages of convolution and Rectified Linear Unit (ReLU) layers to approximate computation (the PCG solver) in the Eulerian fluid simulation. The inputs of the CNN are the divergence of the velocity field, denoted as $\nabla \cdot\ \vec{u}^{*}_{t}$, and the geometry field, denoted as $g_{t-1}$. The output of the CNN is the pressure field, denoted as $\hat{p}_{t}$. The mapping $f_{conv}$ from input to output by this CNN model can be represented as follows:
\begin{equation}
\label{equ:CNN_objective_function}
    \hat{p}_{t} = f_{conv}\Big(\nabla\cdot\ \vec{u}^{*}_{t}, g_{t-1}; W\Big),
\end{equation}
where $W$ is the CNN model parameters. The predicted $\hat{p}_{t}$ is used to update velocity $\vec{u_t}$ in Equation ~\ref{equ:NS_momentum}.
In our study, this CNN (named as the Tompson's model) is used as input in Smart-fluidnet to generate new neural networks for online fluid simulation.

The objective function of the Tompson's model, i.e., $DivNorm$, is the sum of weighted L-2 norm of the divergence of the predicted velocity $\vec{u_t}$ over all fluid cells (mesh volumes) in the rendered smoke frame. $DivNorm$ is defined as follows:
\begin{equation}
    DivNorm = \sum_{i}w_i\{\bigtriangledown \cdot \vec{u_t}\}^{2}_{i},
    \label{Equ:Objective_function}
\end{equation}
where $w_i$ is a weighting term for each fluid cell to emphasize the divergence of grids on geometry boundaries, i.e., $w_i = max(1,k-d_i)$. $d_i$ is the distance field. It takes the value 0 for solid cells or the minimum Euclidean distance to the nearest solid cell for fluid-cells.
\begin{figure}[!t]
  \centering
  \includegraphics[width=0.8\linewidth]{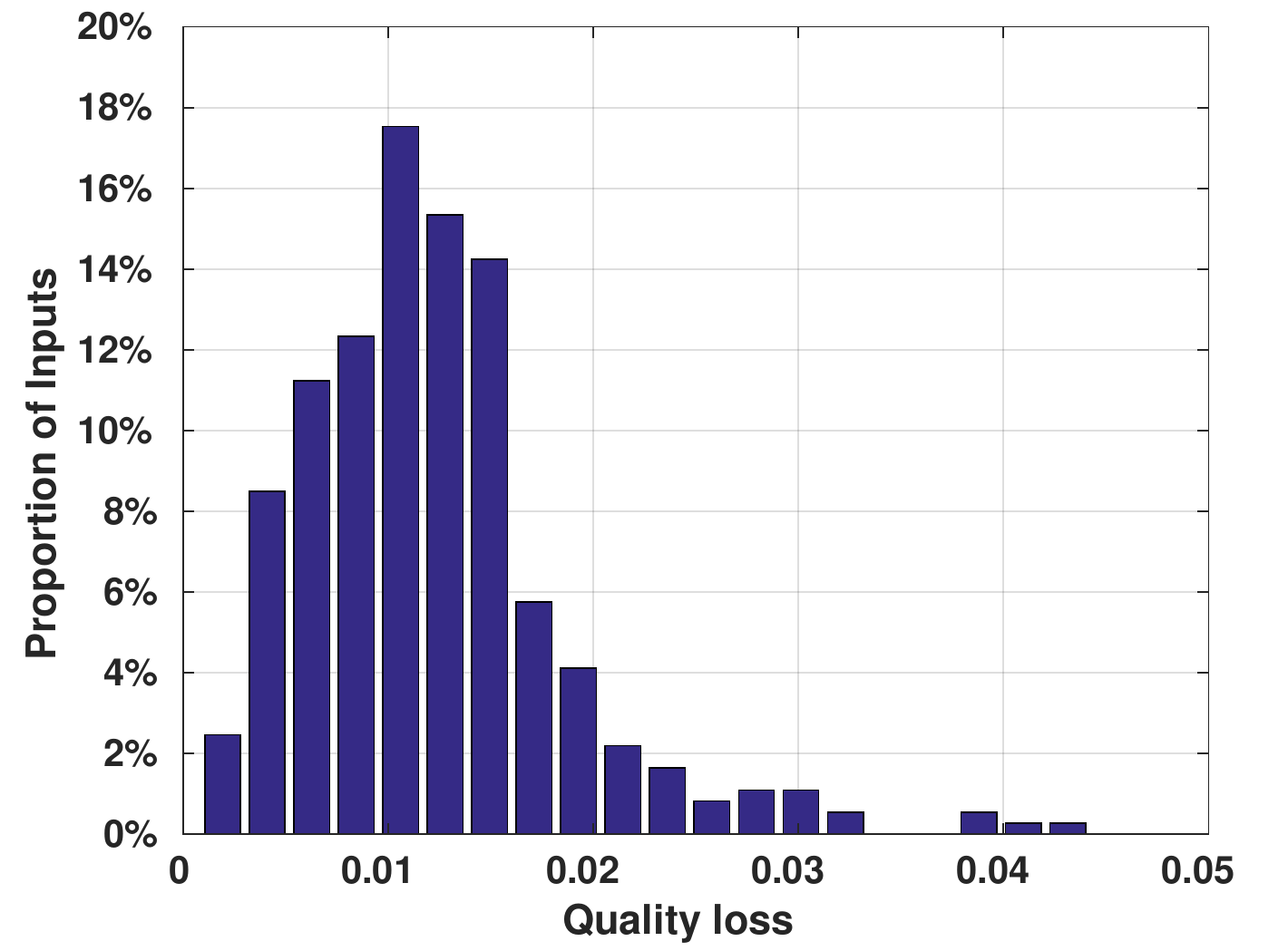}
  \caption{Distribution of quality loss for the Tompson's model with different input problems~\cite{tompson2017accelerating}.}
  \label{fig:Output Quality Variation}
\end{figure}

\vspace{-10pt}
\subsection{Motivation of Our Work}
\begin{table}[!t]
\centering
\caption{Execution time and simulation quality loss of three models for solving the Poisson's equation.}
\label{tab:newton_err_vari}
\begin{tabular}{ l c c }
\toprule
Method      &Execution Time (ms)   & Avg. Quality Loss \\
\midrule   
PCG   & $2.34\times 10^{8}$  & $--$ \\ 
Tompson~\cite{tompson2017accelerating}  & $7.19 \times 10^4$ &$1.3\times 10^{-2}$ \\
Yang~\cite{yang2016data}  & $3.20 \times 10^4$ &$4.9\times 10^{-2}$  \\
\bottomrule
\end{tabular}
\end{table}
The existing work to solve the Poisson's equation includes the PCG solver in mantaflow and two neural network models (Tompson~\cite{tompson2017accelerating} and Yang~\cite{yang2016data}). We study the implications of the three methods on simulation quality and execution time. To study the impact of each model, we evaluate 20,480 different input problems of the fluid simulation and report the average simulation quality loss and execution time. The simulation quality loss $Q_{loss}$ is calculated by comparing the simulation output and the real physical measurements on fluid flow.
Table~\ref{tab:newton_err_vari} and Figure~\ref{fig:Output Quality Variation} show the results.

Table~\ref{tab:newton_err_vari} shows that PCG achieves the highest simulation quality as an exact solution but the worst performance (the longest execution time). On the other hand, the Yang's model performs $10^4\times$ faster than PCG but causes about $10^2\times$ loss in the simulation quality. There is a clear trade-off between simulation quality and performance.

Figure~\ref{fig:Output Quality Variation} shows the distribution of quality loss for various input problems when we use the Tompson's model. The figure reveals that given various input problems, the model generates different simulation quality. For most input problems, the simulation quality loss is between 0.01 and 0.02. Given a user-defined quality requirement (e.g., the quality loss should be less than 0.01), the simulation may not meet the quality requirement for most input problems (e.g., around 65.42\% input problems can not meet user requirement when the requirement is 0.01). We have the same observation for the Yang's model.

The above results reveal that it is imperative to use multiple models to explore the trade-off between performance and simulation quality and {maximize the possibility of reaching the user requirement on the simulation quality for various input problems.}

\begin{figure*}[!t]
  \centering

  \includegraphics[width=1.0\linewidth]{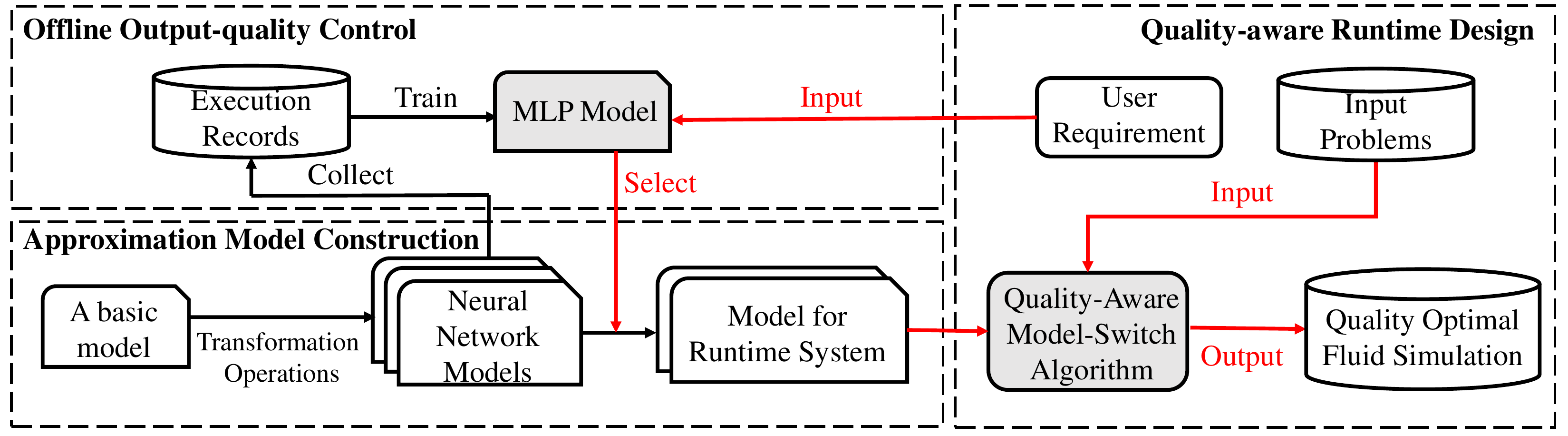}
  \vspace{-20pt}
  \caption{Workflow of the proposed Smart-fluidnet.}
  \label{figure:overview}
\end{figure*}
\section{Overview} 
\label{sec:overview}

Figure~\ref{figure:overview} gives the workflow of Smart-fluidnet. The workflow includes offline and online phases. During the offline phase, Smart-fluidnet takes an existing neural network model as an input and constructs a set of neural network models by model transformation. We introduce four operations (deleting, narrowing, pooling and dropout) to transform the input neural model into multiple neural network models. Then we choose neural network models that are promising for high performance improvement and high quality based on the Pareto optimality analysis.

After model construction, Smart-fluidnet further chooses models based on the user requirement about simulation quality. Given various input problems of the fluid simulation, we introduce an MLP-based model to predict the probability of each model to reach the user requirement on the simulation quality. Considering the possible cost of restarting the simulation when the simulation quality does not meet the user requirement, we choose those models that have a sufficiently high probability to benefit the performance improvement. 
After the above offline phase, we have a handful of neural network models ready for online approximation. 
At runtime, given an input problem of the Eulerian fluid simulation, Smart-fluidnet dynamically switches neural network models with the most promising neural network to meet the user requirement on the simulation quality. The model switching is based on a metric and a linear regression model to predict whether the current model can reach the requirement on the simulation quality at the end of the simulation.

The Smart-fluidnet framework consists of three main modules: approximation model construction (Section~\ref{sec:model_building}), offline output-quality control (Section~\ref{sec:MLP}), and quality-aware runtime design (Section~\ref{sec:lightweight predictor}). Their relationships are depicted in Figure~\ref{figure:overview}. We explain them in detail as follows.

\vspace{-10pt}
\section{Approximate Model Construction}
\label{sec:model_building}

Given an input neural network, we transform it to construct multiple neural network models with different network architectures. 

The new neural network models can be more accurate or more efficient (i.e., using less execution time) than the input neural network; Having such a mixture of different models provides flexible computation approximation during the online fluid simulation.

To generate more accurate neural network models, the user can use an existing framework such as Auto-Keras to generate and train models. Auto-Keras uses the Bayesian optimization to generate a single model with the best accuracy. We change Auto-keras to generate and train five models with the better accuracy. We generate five models, because generating more than five highly accurate models often causes more than five models to be selected after applying MLP (Section~\ref{sec:MLP}), which causes large runtime overhead when making the decision on switching models; While generating less than five models will lead to insufficient candidates after the model selection (Section~\ref{sec:MLP}).

Besides the above, we also aim to generate less accurate but faster models.
We introduce new transformation operations into Auto-Keras to simplify the neural network architecture, which will shorten the execution time. 
We describe our transformation operations as follows. 

\noindent\textbf{Operation 1}: deleting a layer of the neural network. This operation is denoted as $shallow(G, L)$, where $G$ is the neural network graph of the input neural network and $L$ is the layer to be deleted. This operation not only shortens the depth of neural network but also reduces memory consumption,  thus makes the fluid simulation time shorter. 

\noindent\textbf{Operation 2}: narrowing a layer of the neural network. This operation reduces neurons in an intermediate layer. This operation is denoted as $narrow(G, L, r)$, where $r$ is the number of neurons to be reduced at the layer $L$, and $L$ can be either a fully-connected layer or a convolutional layer. 

\noindent\textbf{Operation 3}: pooling. This operation, denoted as $pooling(G, L, m)$, downsamples a layer $L$ using a pooling matrix $m$. We can apply two pooling strategies, i.e., $max pooling$ and $average pooling$.
A special case of $m$ is a $2 \times 2$ matrix which can discard 75\% neurons in the intermediate layers.

\noindent\textbf{Operation 4}: dropout. This operation, denoted as $dropout(G, L, p)$, drops neurons at a layer $L$ with a given probability $p$. 
It offers a more flexible way to reduce the number of neurons, compared with the second operation by controlling the value of $p$. 
This operation is useful to increase the generalization capability of the model.

\begin{figure}[!t]
  \centering
  \includegraphics[width=0.8\linewidth]{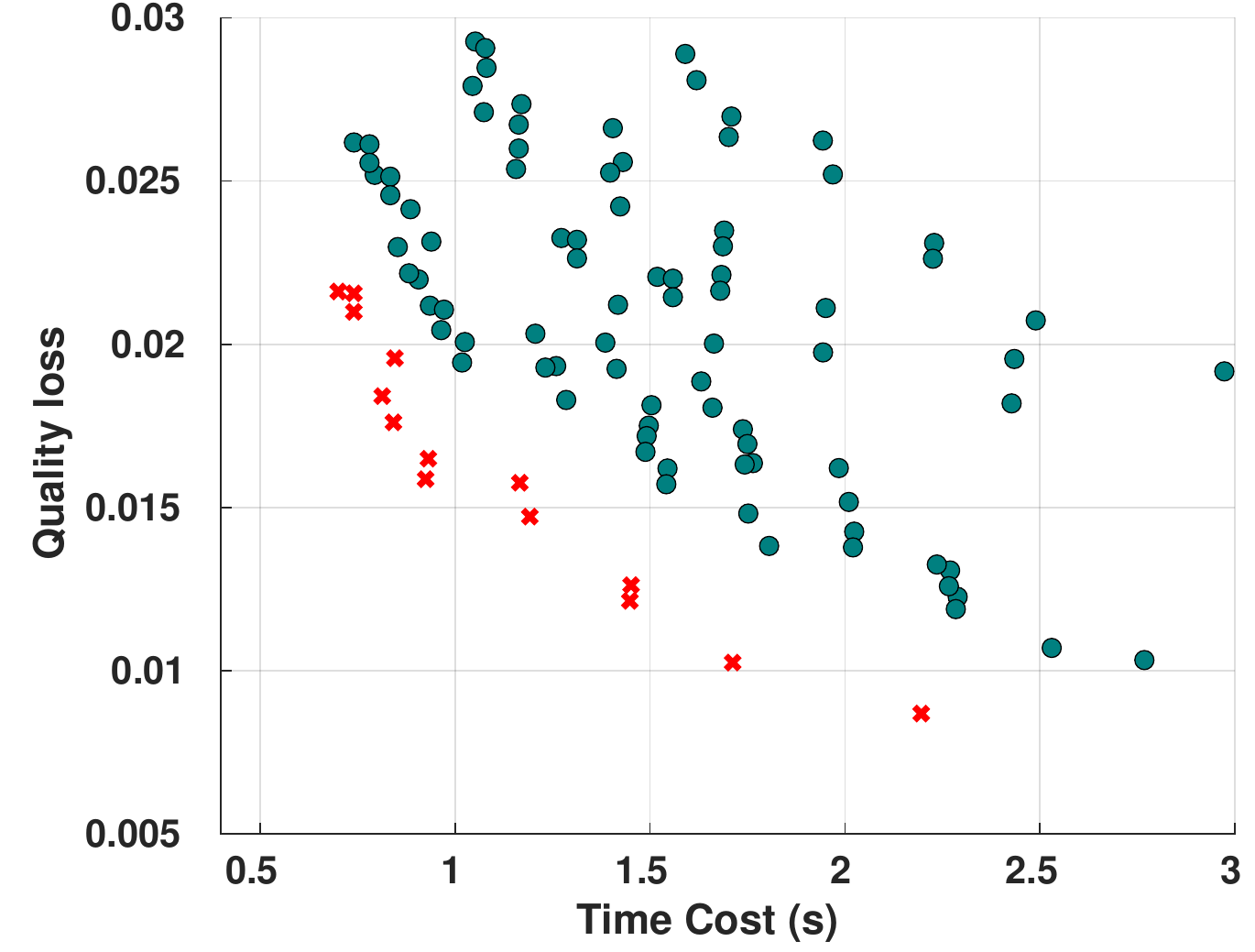}
  \vspace{-10pt}
  \caption{Scatter plot of quality loss and time cost of different neural network models.}
  \label{fig:model_construction_results}
  \vspace{-20pt}
\end{figure}

Based on the above four operations, we transform the input neural network into new ones.
Given an input neural network, we first apply a $shallow(G, L)$ operation on each of the intermediate layers. We do not apply the operation more than twice in the input model, and hence do not delete more than one intermediate layers. 
After applying $shallow(G, L)$, we generate five new models.

Second, we apply $narrow(G, L, r)$ operation on the five new models. A big value of $r$ means a large number of neurons will be reduced. Based on our experimental results, the simulation quality loss can be large (more than 20\%) for most of the input problems, if $r > \frac{|L|}{2}$, where $|L|$ is the total number of neurons. Therefore, we empirically use $r = \frac{|L|}{10}$. 
For each new model, we randomly choose $r$ neurons to apply the $narrow(G, L, r)$ operation; Furthermore, for each new model, we apply $narrow(G, L, r)$ ten times, each of which generates a new model. In our case, in total, we have 55 new models (five new models after applying $shallow(G, L)$ and 50 more after applying $narrow(G, L, r)$).

Third, for the 55 new models, we apply $pooling(G, L, m)$ operations. In particular, for each new model, we randomly replace any of two neighbor-neurons with a new neuron using max pooling. The total number of neurons to be replaced is constrained to be 10\% of the total neurons. After applying $pooling(G, L, m)$, we have 55 more new models (110 models in total). 

At last, to enrich our neuron network models, we randomly select 18 out of the 110 models to apply the $dropout(G, L, p)$ operation. In particular, in each of the 18 models, we randomly drop out neurons. The total number of neurons to drop is limited to 10\% of the total neurons. After applying $dropout(G, L, p)$, we have 18 more new models. In total, we have 128 models.

We apply the four operations in the above order, because the operation that tends to reduce more neurons than other operations will be performed earlier. This method allows us to efficiently generate new models. Using a different order can take longer time to generate models or be prone to generate less accurate models.

After the above model generation and in combination with the accurate models generated by Auto-Keras (five models), we have 133 models in total. Afterwards, we use \textit{Pareto optimality} to reduce the number of models for online approximation. We select models that have the lowest time cost, the lowest quality loss, or both.

To explain the idea of Parento optimality, we use Figure~\ref{fig:model_construction_results} to illustrate it. Figure~\ref{fig:model_construction_results} shows the result of our model selection approach. In this figure, each point represents a model; The red points are those selected model for further analysis (Section~\ref{sec:MLP}); The green points are discarded models. 
In Figure~\ref{fig:model_construction_results}, the quality loss and execution time for each model are collected during the model construction. This is a common method to collect the model information in AutoML work~\cite{Sparks:2015:AMS:2806777.2806945, Li:2018:ETM:3187009.3177737}. Section~\ref{sec:evaluation} gives details on the hardware platform and input datasets to collect the results in Figure~\ref{fig:model_construction_results}.
We can observe that those models located in the leftmost part of Figure~\ref{fig:model_construction_results} either have the lowest time cost, the lowest quality loss, or both. Those models (14 models) are selected based on the Pareto optimality method (we name them ``model candidates'' in the later discussion). 

\noindent\textbf{Sensitivity Study.}
The above process of constructing neural networks involves a couple of parameters. We summarize them as follows and change those parameters to study their impact on the simulation quality. In our study, we use 100 input problems. Using the simulation quality of PCG as the baseline, we calculate the average quality loss of all input problems when using the Tompson's model. This average quality loss is used as the user requirement for quality loss in our sensitivity study.

(1) The number of layers to prune (Operation 1). Our current method prunes one layer at most. For sensitivity study, We prune more than one layer, but find that it leads to a large quality violation (20\% quality loss on average), which is not good.

(2) The percentage of neurons to apply pooling (Operation 3). Our current method 
applies pooling to 10\% of total neurons. We set the percentage of neurons to apply pooling as 5\%,  20\% and 30\%, and evaluate the impact of this parameter on the simulation quality. Our experiments show that using 20\% and 30\%, the fluid simulation has a large quality violation (35\% and 50\% quality loss on average, respectively), which is not good. However, using 5\%, the fluid simulation has the similar quality loss on average as using 10\%. In addition, since using 10\% can lead to better performance than 5\%, we choose 10\% in our study.

(3) The dropout rate (Operation 4). Our current method drops out 10\% of total neurons. We also try 5\% and 15\% as the dropout rate.
Our experiments show that $5\%$ and $10\%$ outperform $15\%$  in terms of the simulation quality loss.
In particular, the quality losses with 5\% and 10\% as the dropout rate are $0.156$ and $0.164$ respectively, while the quality loss with 15\% as the dropout rate is higher ($0.239$). Since 5\% and 10\% has the similar quality loss and using 10\% leads to less execution time, we adopt 10\% as our dropout rate.

(4) The number of neural network models to apply the dropout operation. Our current method chooses 18 models. Our study reveals that choosing 15-20 models are enough for the Pareto optimality and MLP (see Section~\ref{sec:MLP}) to generate 2-5 models. Using less than 15 models to apply the dropout operation, however, we could generate no qualified model after applying MLP. Using more than 20 models, we could have more than 5 qualified models after applying MLP. Having more than 5 models means that the runtime system may suffer from large runtime overhead for selecting a model to use. As a result, we choose 18 (in between 15 and 20) as our parameter.

\section{Offline Output-Quality Control}
\label{sec:MLP}
In many fluid simulation cases, users can have specific requirement on the simulation quality and execution time.
We use the notation, $U(q, t)$, to represent the user requirement, where $q$ and $t$ are the user requirement on the quality loss and execution time respectively. The final quality loss and execution time of the fluid simulation should be less than $q$ and $t$, respectively. The quality control should be aware of the \textit{success rate} of neural network models, namely the ratio of those input problems with which the fluid simulation can reach the simulation quality and time requirement to the total number of input problems.

\begin{figure}[!t]
\centering
    \includegraphics[width=\linewidth, ]{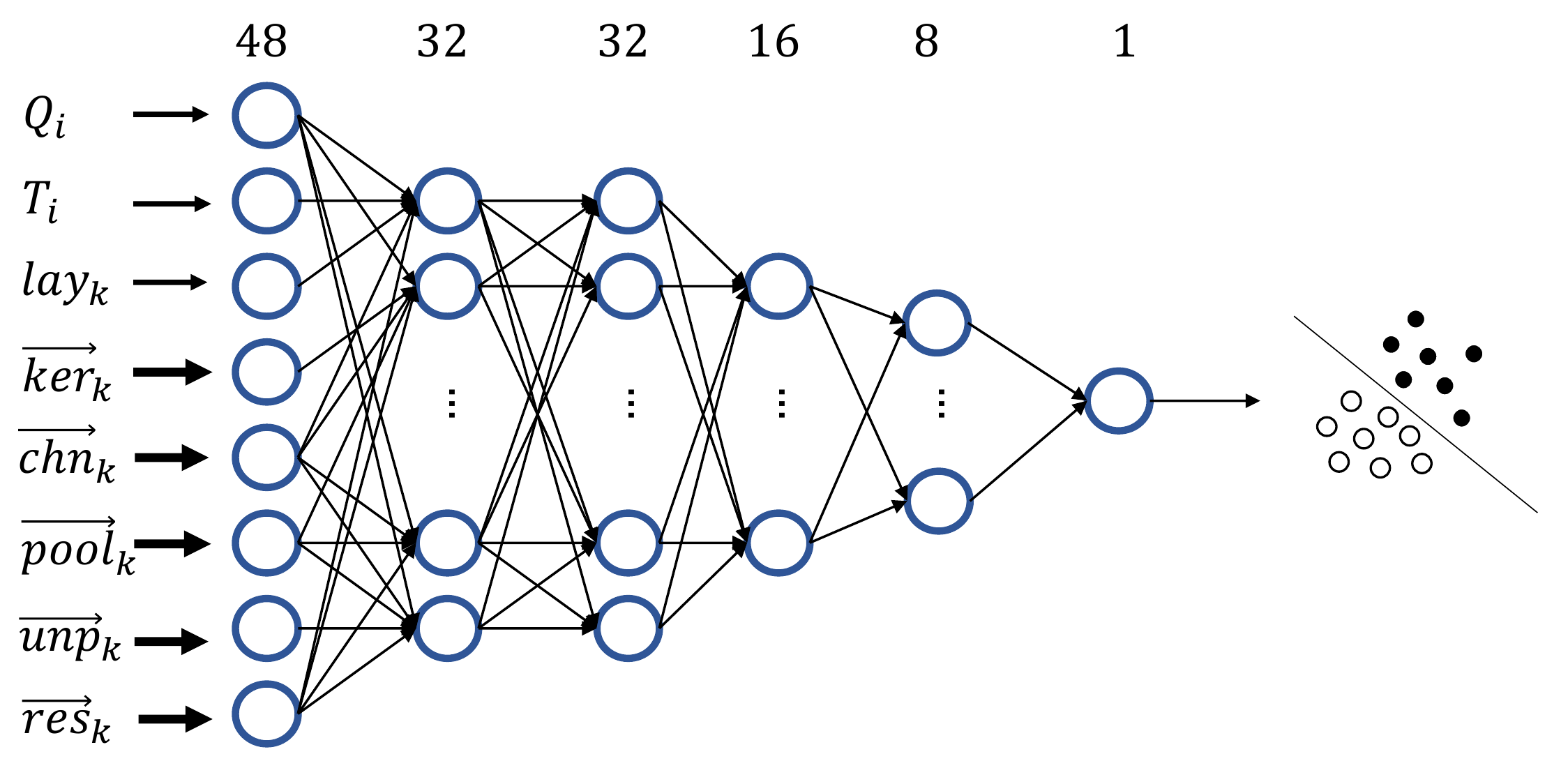}
        \caption{The network architecture of our MLP model.}   
  \label{fig:MLP_model}
  \vspace{-10pt}
\end{figure}

The awareness of the success rate can be developed from statistical knowledge by executing neural network models on various input problems. 
In this section, we design a non-linear MLP model to develop such awareness. 
In the following subsections, we first introduce the construction of training samples for training the MLP model and then give details on how to construct and apply the MLP. 

\subsection{Construction of Training Samples}
\label{sec:mlp_constructing_samples}
We collect execution records (i.e., simulation quality and execution time) for the 14 neural network models after the construction of those models (Section~\ref{sec:model_building}). 
Based on the execution records, we generate training samples to train the MLP model.

\noindent\textbf{Collection of Execution Records.}
For each of the 14 neural network models, we get $N$ execution records by running $N$ input problems ($N=20,480$ input problems in this paper). Each of the $N$ execution records includes the simulation quality $q_n^k$ and execution time $t_n^k$, where $n \in [1, N]$ and $k$ refers to a neural network $NN_k$ from the 14 neural network models. Each execution record is represented as $ER_n^k$. Such execution record is collected during the model construction (Section~\ref{sec:model_building}). 
Note that our model construction process, similar to other AutoML work~\cite{Sparks:2015:AMS:2806777.2806945, Li:2018:ETM:3187009.3177737}, includes not only changing model architecture but also training models. Hence we can collect execution records during the model construction. 

\noindent\textbf{Sample Generation.}
Given $N$ execution records, we can generate samples to train the MLP. Each sample is represented with a feature vector using Equation~\ref{eq:training_sample_for_mlp}. 

\begin{equation}
\label{eq:training_sample_for_mlp}
F = \Big(q, t, l_k,  \overrightarrow{ker}_k, \overrightarrow{chn}_k, \overrightarrow{pool}_k, \overrightarrow{unp}_k, \overrightarrow{res}_k\Big)
\end{equation}

In Equation~\ref{eq:training_sample_for_mlp}, the feature vector $F$ includes quality and execution time requirements ($q$ and $t$), and architecture information for the neural network $NN_k$ (i.e., $l_k$, $\overrightarrow{ker}_k$, $\overrightarrow{chn}_k$, $\overrightarrow{pool}_k$, $\overrightarrow{unp}_k$, and $\overrightarrow{res}_k$, representing the number of layers, kernel sizes,  channel number,  pooling size, unpooling size,  and residual connection of each layer respectively). Each of the last five architecture information is a vector composed of nine components that indicate properties of each layer of the neural network $NN_k$. Therefore, each input feature vector has $3+5*9=48$ components in total.

Given a neural network $NN_k$, we generate a sample by randomly picking up a user requirement ($q$ and $t$), and then use Equation~\ref{eq:training_sample_for_mlp} to build the feature vector of the sample. Based on the $N$ execution records and the user requirement, we calculate that how many of $N$ execution records meet the user requirement. The ratio of those execution records to $N$, denoted as $r_{k,q,t}$, is the label of the sample. By choosing different combinations of $q$ and $t$, we can generate as many samples as possible.

\subsection{MLP Model Construction and Loss Function}
Given a neural network $NN_k$ and user requirement $q$ and $t$, we can build a feature vector $F$ using Equation~\ref{eq:training_sample_for_mlp}. Our MLP model (see Equation~\ref{eq:mlp}) takes such a feature vector as input and generates an output $\hat{r}_{k,q,t}$, which is a floating point number indicating the probability that $NN_k$ meets the user requirement for any input problem.
\begin{equation}
\label{eq:mlp}
    \hat{r}_{k,q,t} = f_{MLP}\big(F_{k,q,t}\big)
\end{equation}

Figure~\ref{fig:MLP_model} shows the network topology of our MLP. It includes six hidden layers and a 48-neuron input layer. The numbers of neurons in the six hidden layers are 32, 32, 16, 16, 8 and 8 respectively. All the neurons in the hidden layers use ReLU as activation to increase the non-linearity of the model. The neurons in the last hidden layer uses a sigmoid function as activation. 
Using the samples constructed in Section~\ref{sec:mlp_constructing_samples} to train the MLP, we aim to minimize the loss between the model output $\hat{r}_{k,q,t}$ and the ground truth label $r_{k,q,t}$.

\begin{figure}[!t]
    \centering
    \includegraphics[width=0.75\linewidth, ]{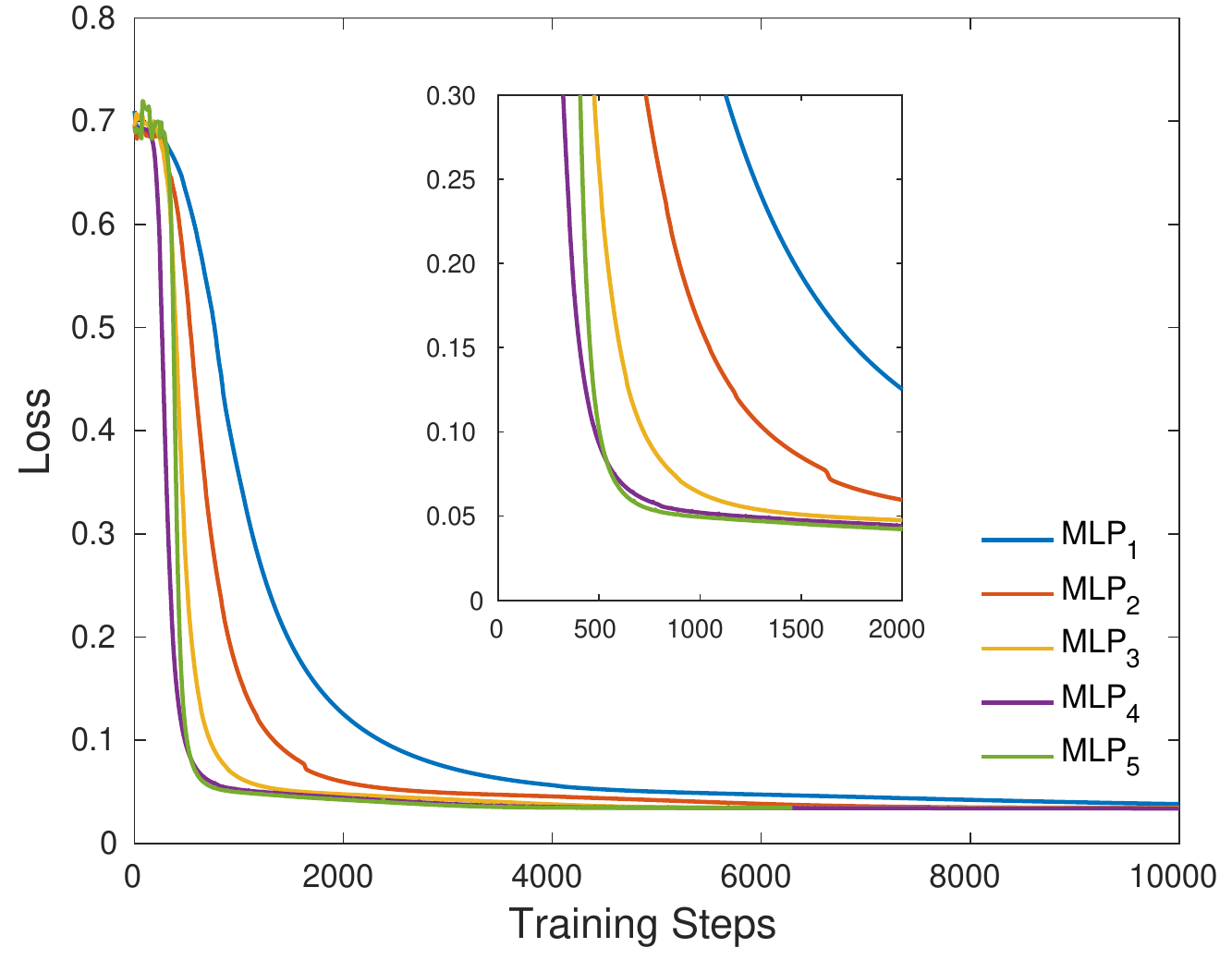}
    \vspace{-10pt}
       \caption{Training losses of five MLPs. 
        } 
\vspace{-15pt}
  \label{fig:mlp_loss}
\end{figure}

\noindent\textbf{Alternative MLP Topologies.} Besides the above MLP topology, we try four alternative MLP topologies, in order to find one for best accuracy. {Among the four MLPs, two of them are deeper than our current MLP, while the other two are shallower.} When building an alternative MLP, we follow the rule that in the topology of a neural network, a deeper layer generally has less number of neurons than shallower ones (i.e., {the number of neurons gradually decreases across layers}).
Many discriminative neural networks, such as Alex-Net\cite{krizhevsky2012imagenet},  VGG-Net\cite{simonyan2014very}, are constructed, following this rule. The architectures of the four models plus our current MLP model are briefly described as follows: 
\begin{itemize}
    \item $MLP_1$ has 4 layers with 48, 32, 16 and 1 neurons;
    \item $MLP_2$ has 5 layers with 48, 32, 16, 8 and 1 neurons;
    \item $MLP_3$ (our current MLP model) has 6 layers with 48, 32, 32, 16, 8 and 1 neurons;
    \item $MLP_4$ has 7 layers with 48, 64, 32, 32, 16, 8 and 1 neurons;
    \item $MLP_5$ has 8 layers with 48, 64, 64, 32, 32, 16, 8 and 1 neurons.
\end{itemize}

Figure~\ref{fig:mlp_loss} presents the training loss curves of the above five MLPs ($MLP_3$ is our current MLP model). We find that the convergence speed of $MLP_3$ is faster than those of $MLP_1$ and $MLP_2$, and offers lower training loss (i.e., higher prediction accuracy). Compared with $MLP_3$ model, $MLP_4$ and $MLP_5$ do not have significant advantages {in terms of convergence speed and loss}, although they have deeper topologies. Hence, $MLP_3$ exhibits a balanced trade-off between prediction accuracy and model size, and is thus chosen as our MLP model in this paper.

\subsection{Usage of MLP}
Given a user-specified simulation quality ($q$) and time cost ($t$), we use MLP to calculate $\hat{r}_{k,q,t}$ for a given neural network model $NN_k$. A larger value of $\hat{r}_{k,q,t}$ represents a higher success rate. In other words, it is highly possible that $NN_k$ can meet the simulation quality and time requirement on an input problem.

Given the user-specified simulation quality ($q$) and time requirement ($t$), the neural network $NN_k$ and MLP prediction result ($\hat{r}_{k,q,t}$), we use the following method to decide if $NN_k$ should be selected for      {the runtime system for the fluid simulation}. In particular, considering the probability that the user requirement on the simulation quality is violated and the user has to re-run the simulation without using any neural network, the simulation time is calculated based on Equation~\ref{eq:simulation_time}. 
 \begin{equation}
 \label{eq:simulation_time}
     T_{total}  = \hat{r}_{k,q,t} \times T_{M_{k}} + (1- \hat{r}_{k,q,t}) \times T',
 \end{equation}
where $T'$ is the execution time without using any neural network and $T_{NN_{k}}$ is the execution time using the neural network $NN_k$. We compare $T_{total}$ with the user requirement on the execution time $t$. Only those neural networks that have $T_{total}$ less than $t$ is selected.

The above selection method considers the impact of violating the simulation quality requirement on the simulation time, and ensure that if $NN_k$ is repeatedly employed for many input problems, there is performance benefit.

\section{Quality-Aware Runtime Design}
\label{sec:lightweight predictor}

After applying MLP, multiple neural networks are selected. We use a runtime technique to schedule those neural networks to {optimize performance and meet the simulation quality requirement}. In order to determine which neural network should be used at runtime, we need to evaluate the model being used in terms of the final quality loss, and switch to a suitable one if necessary. However, without running the simulation to completion, we cannot know the final quality loss. Thus we construct a metric called $CumDivNorm$ (defined in Equation~\ref{equ:cumdivnorm}), which is used to set up a bridge between $DivNorm$ (see Equation~\ref{Equ:Objective_function}) measurable at runtime and the final simulation quality loss $Q_{loss}$  
(Section ~\ref{Prediction of Quality Loss}). 
Based on $CumDivNorm$ and the predicted final quality loss, we introduce a quality-aware model-switch algorithm (Section ~\ref{Quality-Aware Model-Switch Algorithm}) to select the best neural network to accelerate the fluid simulation.
\subsection{Prediction of Simulation Quality Loss}
\label{Prediction of Quality Loss}

The objective function $DivNorm$ provides a goal that our neural network aims to achieve. Using $DivNorm$, we can know how the neural work performs in terms of prediction accuracy. However, there is a missing link between the prediction accuracy of the neural network and the final simulation quality loss $Q_{loss}$.

\noindent\textbf{\emph{CumDivNorm}: A Metric for Runtime Quality Control.} 
To explore the relationship between $DivNorm$ and final simulation quality $Q_{loss}$, we calculate $CumDivNorm$ (i.e., the accumulation of $DivNorm$) and $Q_{loss}$ at \textit{each simulation time step} (denoted as $Q_{loss}^{ts}$). The accumulation of $DivNorm$ over $n$ time steps is defined in Equation~\ref{equ:cumdivnorm}.

\begin{equation}
\label{equ:cumdivnorm}
    CumDivNorm = \sum_{i=1}^nDivNorm_{i}.
\end{equation}

\begin{figure}[!t]
    \centering
    \includegraphics[width=\linewidth]{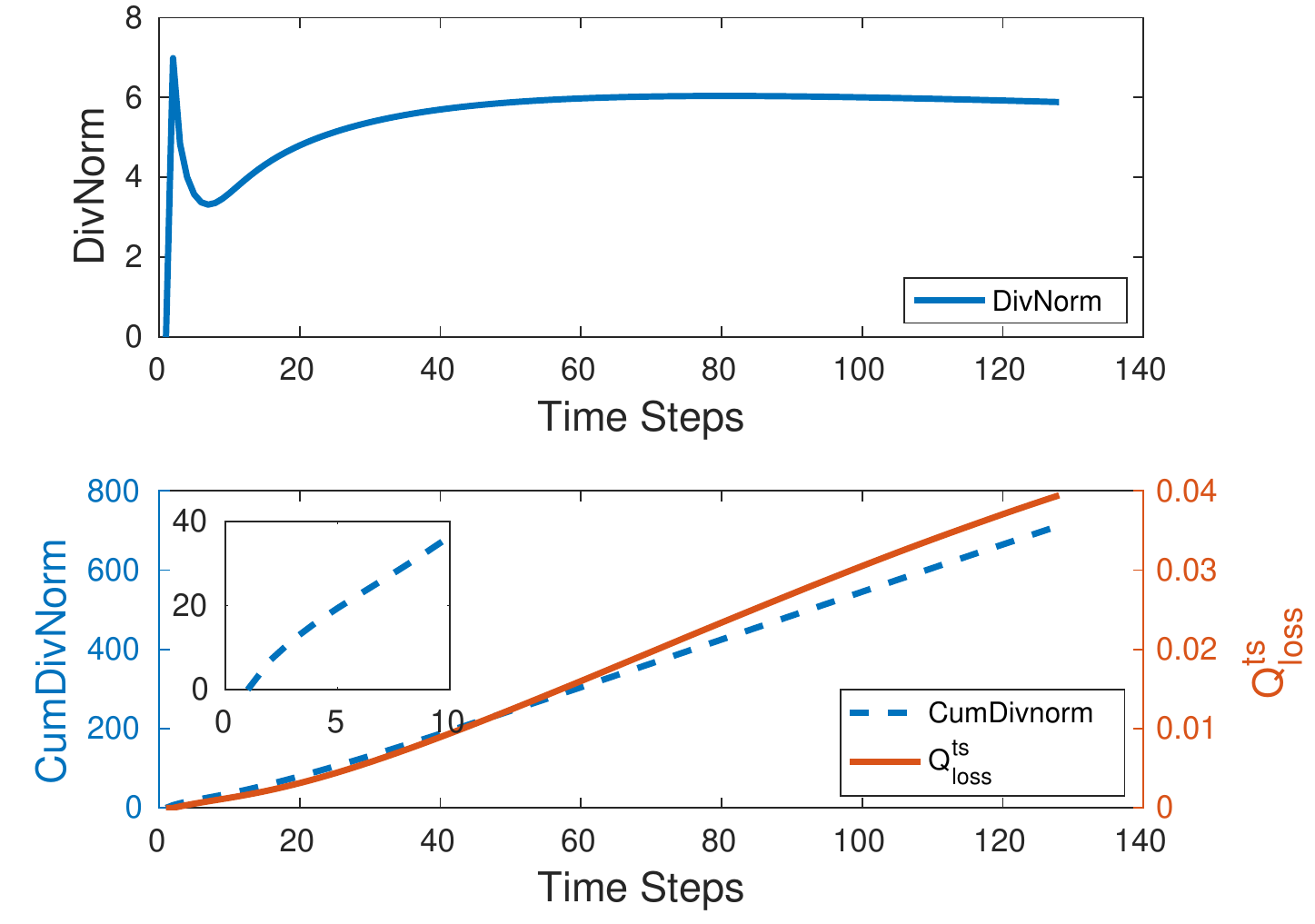}
    \vspace{-20pt}
        \caption{Relationship between $CumDivNorm$ and $Q_{loss}^{ts}$.} 
  \label{fig:sec6_1}
  \vspace{-15pt}
\end{figure}

In order to observe how these varibles are correlated, {Figure~\ref{fig:sec6_1} depicts how $DivNorm$, $CumDivNorm$, and $Q_{loss}$ vary across all time steps of the fluid simulation using an input problem with the grid size 1028*1028. We have the following observations. These observations are valid for other input problems as well.}

\begin{itemize}
    \item \textbf{Observation 1}: $DivNorm$ dramatically increases at the first few time steps and then gradually converges to a stable value;
    \item \textbf{Observation 2}: {$Q_{loss}^{ts}$ and $CumDivNorm$ have similar increasing tendency (except the first few time steps).}
\end{itemize}

The above observations indicate that  $CumDivNorm$ and $Q_{loss}^{ts}$ calculated at each time step are correlated. In order to quantify the relationship between $CumDivNorm$ and $Q_{loss}^{ts}$ at each time step, we use
the Pearson's product moment correlation coefficient ($rp$) \cite{nefzger1957needless} and the Spearman's rank correlation coefficient ($rs$) \cite{henrysson1971gathering} to statistically reveal the correlation between the two variables.

The two coefficients are defined as follows.
Given two input vectors $x$ and $y$ (the vector length is $n$), the calculation of $rp$ and $rs$ to quantify the correlation between $x$ and $y$ is formulated as follows:
\begin{equation}
\label{eq:rp}
rp = \frac{{}\sum_{i=1}^{n} (x_i - \overline{x})(y_i - \overline{y})}
{\sqrt{\sum_{i=1}^{n} (x_i - \overline{x})^2(y_i - \overline{y})^2}},
\end{equation}
\begin{equation}
\label{eq:rs}
rs = 1- {\frac {6 \sum d_i^2}{n(n^2 - 1)}}.
\end{equation}
where $x_i$ and $y_i$ are the values of $x$ and $y$ for the $i{\text{-th}}$ component, and $d$ counts the pairwise disagreement (i.e., $x_i$ is not equal to $y_i$) between the two vectors (see \cite{best1975algorithm} for a more detailed description). 
In general, the coefficients that belong to [0.10-0.29] represent weak association, (0.29-0.49] represent medium association, and above 0.49 represent strong association \cite{chok2010pearson}. 

In our study, we use 20,480 input problems, each of which will have 128 simulation time steps.
The  $CumDivNorm$ and $Q_{loss}^{ts}$ of each time step will be calculated to build the two input vectors. 
Using Equations~\ref{eq:rp} and~\ref{eq:rs}, we have $rp=0.61$ and $rs=0.79$, which indicates a strong correlation.

Based on the above discussion, it is possible to use $CumDivNorm$ to predict $Q_{loss}^{ts}$ in the final time step (i.e., the final quality loss $Q_{loss}$). Note that we cannot calculate $Q_{loss}^{ts}$ at runtime, because it involves PCG, which is too expensive. In the following discussion, we first discuss how to predict $CumDivNorm$ in the final time step ($CumDivNorm^{final}$), and then we discuss how to predict $Q_{loss}$ based on the predicted $CumDivNorm^{final}$. 

\noindent\textbf{Predicting \emph{CumDivNorm\textsuperscript{final}}}. 
We introduce a lightweight approach to predict $CumDivNorm^{final}$. Our prediction approach is based on Figure~\ref{fig:sec6_1}. 
{In this figure, $CumDivNorm$ quickly grows at first, and then the growth rate remains stable. This trend is general across all 20,480 input problems we test.}

The stable growth rate of $CumDivNorm$ makes it possible to predict $CumDivNorm^{final}$ in the middle of the fluid simulation. In particular, we use five time steps to build a linear regression model ($f_{k}(x) = ax + b$), by the least square method, where $x$ is the time step and $f_{k}(x)$ is the predicted $CumDivNorm$. Note that the data used to build the linear regression model must be collected \textit{after} the growth rate becomes stable. Hence, we skip the first five time steps and build the regression model after each five steps. Also, in each five time steps (a check interval) to build the model, we skip the first two to make sure the trend is stable and only use the remaining three to build the model.
The above model-building process happens every five time steps, and the model is used to predict and \textit{check} $Q_{loss}$ (see the following discussion). Hence, we fix the check interval to be five time steps in the rest of the paper. In Section~\ref{sec:eva_check_interval}, we study the impact of the check interval on the simulation quality.
 
\begin{algorithm}[!t]
\small
\caption{The quality-aware model-switch runtime algorithm} 
\label{alg:online_algorithm}
\begin{algorithmic}[1]
\REQUIRE The user requirement $U(q, t)$.
\STATE Choose a neural network model $M_k$ with the highest success rate according to MLP. 
\WHILE {$t$ does not reach the final time step}
\STATE {Send $M_k$ to predict the final simulation quality.}  
\STATE \textbf{Prediction of quality loss:}
\STATE 1) {Build a linear regression model with $DivNorm$ values measured in the last five time steps;}
\STATE 2) Predict $CumDivNorm^{final}$ by the linear regression model;
\STATE 3) Predict $Q_{loss}'$ of the current neural network model by the KNN algorithm; 
\STATE  \textbf{Model Switch:}
\IF{$Q_{loss}'$ is close to $q$}
\STATE Continue using the current neural network model for L steps; 
\ELSIF{$Q_{loss}'$ less than $q$ }
\STATE Switch to a faster (less accurate) neural network model;
\ELSIF {$Q_{loss}'$ is larger than $q$}
\STATE Switch to a slower (more accurate) neural network model;
\ELSIF {Cannot find any model}
\STATE Restart by the PCG method;
\ENDIF
\STATE $t \gets t+L$ ; //L is the check interval. 
\ENDWHILE 
\RETURN{0}
\end{algorithmic}
\end{algorithm}
\noindent\textbf{Predicting \emph{Q\textsubscript{loss}} based on \emph{CumDivNorm\textsuperscript{final}}.}
We use a method based on the k-nearest neighbor (KNN) algorithm to predict $Q_{loss}$. Our method includes offline and online phases. During the offline phase, we test the neural network models selected by MLP with 128 small input problems. For each test, we collect a pair of data ($CumDivNorm^{final}$, $Q_{loss}$) and put them into a historical database. The offline phase is fast, because we use small input problems. During the online phase, at a time interval, to predict $Q_{loss}$, we check $CumDivNorm^{final}$ in the database and find $k$ pairs whose $CumDivNorm^{final}$ are the closest to the predicted $CumDivNorm^{final}$ in the current time interval. We use the average of $Q_{loss}$ in the $k$ pairs as the predicted $Q_{loss}$ in the current time interval. In our evaluation, we use different values for $k$, but find that $k \in [4, 6] $ is usually sufficient to give accurate prediction, hence we choose $k=4$ to reduce runtime overhead.  

For example, assuming that the predicted $CumDivNorm^{final}$ in a time interval is 108. To predict  $Q_{loss}$ for this time interval, we select four pairs from the database, i.e., (101, 0.09), (112, 0.11), (105, 0.10), and (109, 0.11),  
whose $CumDivNorm^{final}$ is closest to the given $CumDivNorm$ (108). 
Then the predicted $Q_{loss}$ for this time interval is 0.1025 $\big((0.09+0.11+0.10+0.11)/4\big)$. We organize all data pairs as a binary search tree, such that finding the four pairs is cheap.

\subsection{Quality-Aware Model-Switch Algorithm}
\label{Quality-Aware Model-Switch Algorithm}

With the ability to predict the simulation quality loss,  we introduce a quality-aware model-switch algorithm. Algorithm~\ref{alg:online_algorithm} depicts this runtime algorithm. After applying MLP, we have several promising neural network models and their probabilities to meet the user-specified requirement. We also know the execution time (i.e., the inference time) of each network model. During the simulation, the neural network with the highest probability to meet user-specified requirement is selected as the first model to approximate computation in the fluid simulation. 
Then we calculate $CumDivNorms$ in the first check interval, build a linear regression model, calculate $CumDivNorm^{final}$ using the regression model, and predict $Q_{loss}$ by the KNN algorithm. After that, we compare the predicted $Q_{loss}$ (annotated with $Q_{loss}'$ in the rest of the paper) with the user requirement $q$. If $Q_{loss}'$ is close to $q$, we predict that the current neural network model can meet the user requirement. The runtime algorithm continues to use the current neural network model. But if $Q_{loss}'$ is larger (or smaller) than $q$, then the runtime algorithm chooses a accurate (or fast) model with better (or worse) accuracy. If all the neural network models cannot meet $q$, we restart the simulation and use the traditional simulation method (i.e., the PCG method). The above model switch process happens periodically (the period is the check interval). We calculate $CumDivNorms$ at the end of every check interval to determine if the model switch is necessary.

\begin{figure}[!t]
    \centering
    \includegraphics[width=0.5\textwidth]{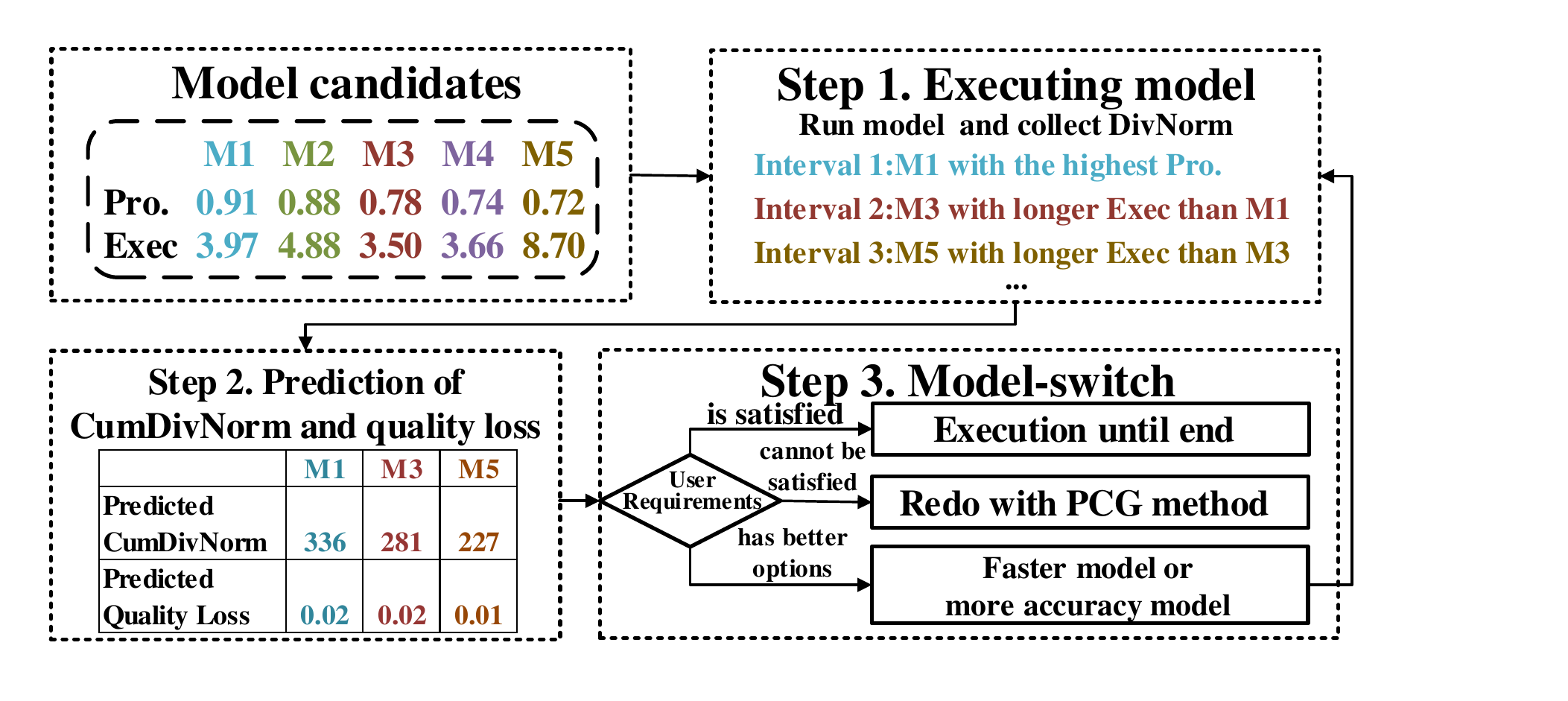}
    \vspace{-10pt}
        \caption{An example to explain our runtime algorithm.} 
  \label{fig:sec6_example}
  \vspace{-10pt}
\end{figure}

\noindent\textbf{An Example.}  
Figure~\ref{fig:sec6_example} gives an example to further explain our runtime algorithm. In this example, we have five neural network models and the user requirements on the simulation quality loss and execution time are 0.013 and 6.64s respectively.

During the offline phase, five neural network models are constructed by the model transformation (Section~\ref{sec:model_building}) and selected by MLP (Section~\ref{sec:MLP}); We record the possibility and execution time of the five neural network models (shown as Step 1 in Figure ~\ref{fig:sec6_example}). At runtime, we use the first neural network ($M1$) which has the highest probability (91\%) to meet the user requirement on the simulation quality. Then we skip the first five steps and fit the values of $DivNorms$ in the first check interval into a linear regression model, and predict that the $CumDivNorm$ closes to 395 at the final time step. Using the KNN method, we predict $Q_{loss}$ as 0.019, which is much larger than the user requirement (0.013). So we switch to a more accurate neural network model, i.e., $M3$ (a model with higher accuracy than $M1$ and the highest probability among remaining neural network models). After using $M3$ for another five time steps (one check interval), we predict $Q_{loss}'$ of $M3$ as 0.015, which is still larger than the user requirement.
So we switch to another more accurate neural network model, i.e., $M5$. We predict $Q_{loss}'$ of $M5$ as 0.013, which can meet the user requirement, we then use $M5$ during the next check interval.

Compared with using a single neural network model, our runtime algorithm introduces additional computation (i.e., predicting $CumDivNorm$ and applying the KNN method). However, the computation of the simple linear regression algorithm to predict $CumDivNorm$ and the traversal of the binary tree to apply the KNN method are lightweight. Such overhead can be easily overweighed by the performance benefit introduced by adaptive neural network-based approximation. We evaluate performance (including the runtime overhead) in Section ~\ref{sec:Analysis_of_Runtime_System}.

\section{Evaluation}
\label{sec:evaluation}
We evaluate our framework to examine its impact on performance and simulation quality of the Eulerian fluid simulation.

\noindent\textbf{Platform.} We conduct all experiments on a high-end server with 24 Intel Xeon E6-2760 v3 CPU cores running at 2.30GHz. The server is equipped with an NVIDIA Titan X (Pascal) GPU. We use cuDNN 5.0 on this GPU to run neural networks.

\noindent\textbf{Fluid Simulation.}
During the fluid simulation, we run 128 time steps (the default number of time steps in mantaflow) for each input problem. To comprehensively evaluate the performance, {we use multiple grid sizes for each input problem}, including 128$*$128, 256$*$256, 512$*$512, 768$*$768, and 1024$*$1024.

\noindent\textbf{Input Datasets.}
We generate training and evaluation datasets by {mantaflow}, which is an open-source framework for the fluid simulation.
The training dataset is used to optimize the parameters of the neural network models and MLP model, while the evaluation dataset is used to evaluate the performance of the online algorithm during the fluid simulation.
Each of the two datasets contains 20,480 input problems. There is no overlapping between the training and test datasets. To generate the total 40,960 problems, we initialize velocity by a pseudo-random turbulent field~\cite{kim2008wavelet}, and generate occupancy grids with the border wall by introducing some objects in the simulation domain. Those objects are from the NTU 3D Model Dataset~\cite{pu2006visual}. 

\noindent\textbf{Neural Networks}. 
We use Auto-Keras~\cite{jin2018efficient} with extension (Section~\ref{sec:model_building}) to search qualified convolutional neural network architectures. Auto-Keras constructs neural networks using Python, but the fluid simulation is implemented in Lua/Torch7. Hence we re-implement and optimize the neural networks with the Torch7 package. 

\begin{figure}[!t]
    \centering
    \includegraphics[width=\linewidth, ]{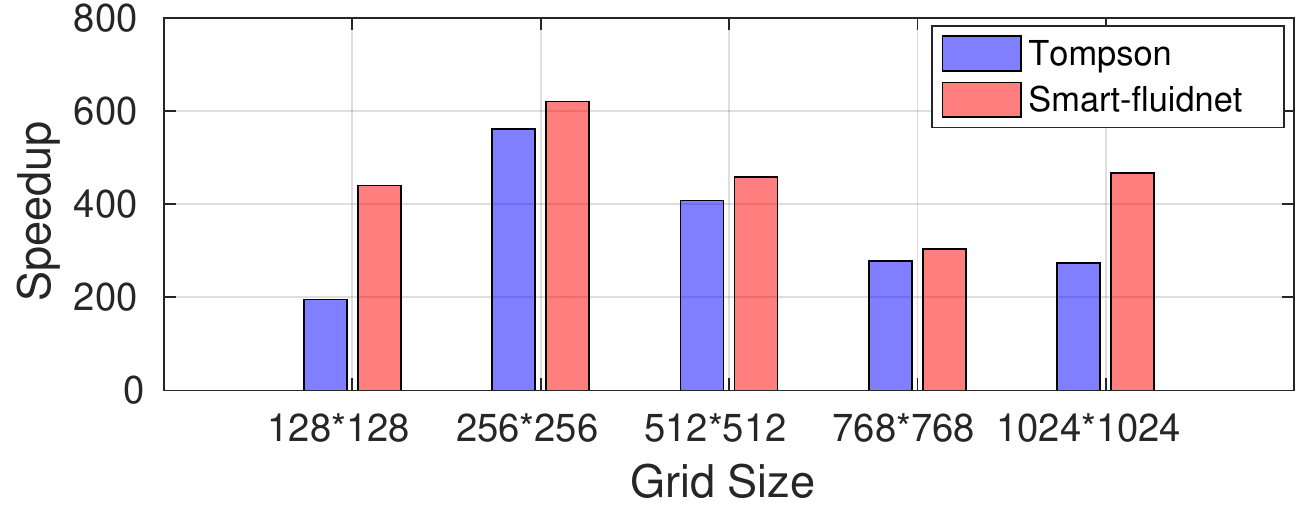}
       \vspace{-18pt} 
        \caption{Performance (execution time) of the Tompson's model and Smart-fluidnet. 
        } 
  \label{fig:Speedup}
\end{figure}

\begin{figure}[!t]
    \centering
    \includegraphics[width=\linewidth, ]{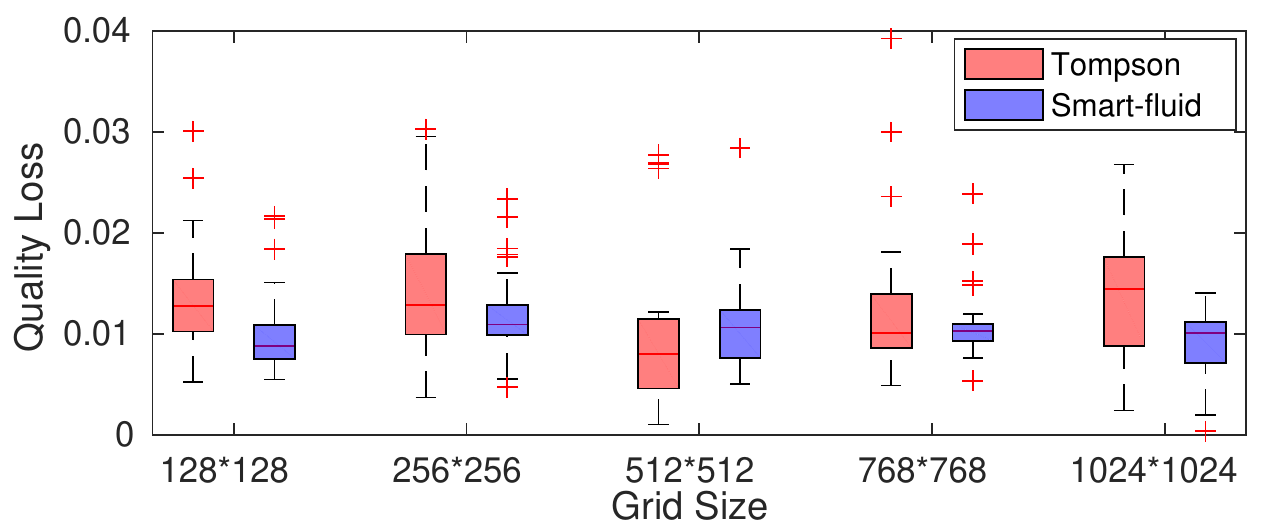}
       \vspace{-18pt}
        \caption{Variation of quality loss with various grid sizes for input problems.} 
  \label{fig:sec7_2}
\end{figure}

\subsection{Model Speedup and Accuracy}
\label{sec:model_speedup_accuracy}
We conduct experiments to measure performance. We take the PCG solver as the baseline method. PCG is the traditional method used in the Eulerian fluid simulation and does not include any neural network. Compared with neural network-based approximation, PCG has the highest simulation quality but the performance is very bad. All performance (execution time) reported in this section is shown as ``speedup'' with respect to the performance of PCG.

Figure~\ref{fig:Speedup} shows the results for Smart-fluidnet and the Tompson's model with different grid sizes. The Tompson's model represents the {state-of-the-art neural network} to accelerate the Eulerian fluid simulation. In all test cases, Smart-fluidnet performs better than the Tompson's model and is $1.46 \times $ better on average.
The largest improvement over the Tompson's model is $2.25 \times$.
Besides the execution time, we also study the simulation quality of the Tompson's model and Smart-fluidnet. We use 20,480 input problems to evaluate each method. We use the simulation quality of PCG as the ground truth and study the quality loss of the Tompson's model and Smart-fluidnet. Since Smart-fluidnet requires the user to specify a requirement on the quality loss, we use the average quality loss of all input problems when using the Tompson's model, as the user requirement (the target).

\begin{table}[!t]
\centering
\caption{Percentage of input problems with which the simulation reaches the requirement on quality.}
\vspace{-8pt}
\label{tab:success rate}
\begin{tabular}{l p{0.8cm}p{0.8cm}p{0.8cm}p{0.8cm}p{1.2cm}}
\hline
Grid size & 128*128 & 256*256 & 512*512&768*768 & 1024*1024 \\
\hline   
Tompson & 68.22\% & 67.16\% &85.27\% &71.06\% & 46.38\%\\
Smart-fluidnet & 88.27\% & 87.14\% & 91.36\%& 86.47\%& 91.05\% \\
\hline
\end{tabular}
\end{table}

Figure~\ref{fig:sec7_2} presents boxplots\footnote{In the boxplots, the boxes are bounded by 25th and 75th percentiles of the quality loss; The central marks of the boxes indicate the median; The `+' markers outside the boxes indicate the extreme outliers~\cite{dawson2011significant}. } to show the results of the quality loss for all input problems with five selected grid sizes.
We draw two observations from Figure~\ref{fig:sec7_2}: (1) The outputs of Smart-fluidnet are closer to the target value than those of the Tompson's model; (2) The variances of Smart-fluidnet are smaller than those of Tompson's model.
These two observations reveal that Smart-fluidnet can give more \textit{consistent simulation quality} than the Tompson's model, which is crucial for dealing with largely diversified input problems.

To further study the consistency of simulation quality with various input problems, we study how many input problems can lead to the simulation with satisfiable simulation quality. 
Table~\ref{tab:success rate} shows the results. Table~\ref{tab:success rate} reveals that Smart-fluidnet leads to a larger percentage of a high-quality simulation than the Tompson's model in all cases. The difference between Smart-fluidnet and Tompson's model is as large as 44.67\% (when the grid size is 1024*1024).

\subsection{Analysis on Runtime System}
\label{sec:Analysis_of_Runtime_System}
In this section, we analyze the effectiveness of our runtime system. Taking the grid size of 1024*1024 as an example, the average quality loss and execution time of the Tompson's model are 0.013 and 6.64 seconds respectively. We take this quality loss and execution time as the target (i.e., the user requirement) of Smart-fluidnet.

\begin{figure}[!t]
    \centering
    \includegraphics[width=\linewidth, ]{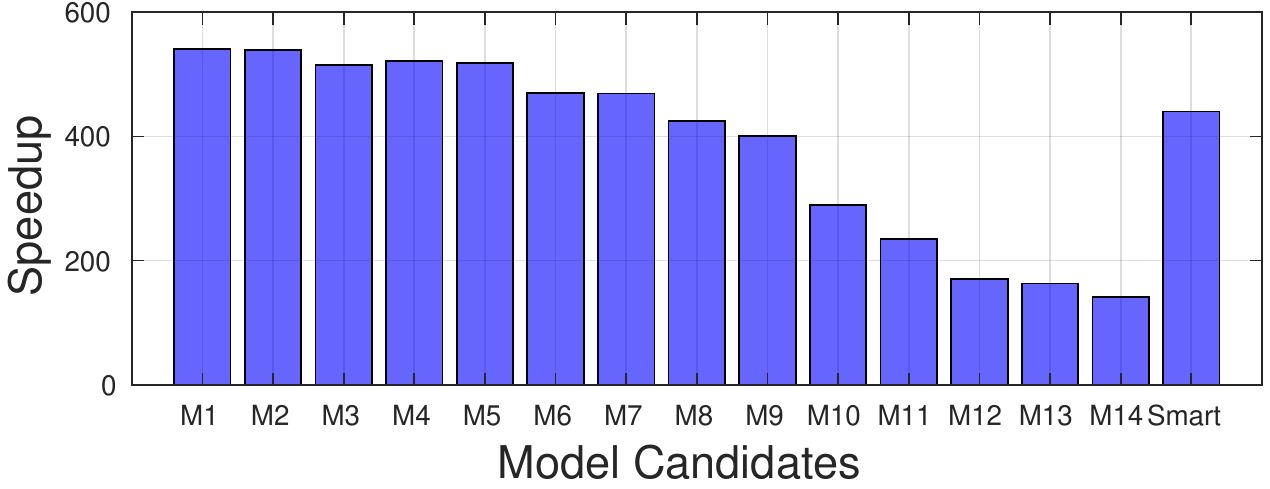}
    \vspace{-15pt}
        \caption{Performance (execution time) for the Tompson's model and Smart-fluidnet.} 
  \label{fig:sec7_4}
\end{figure}

\begin{figure}[!t]
\centering
    \includegraphics[width=\linewidth, ]{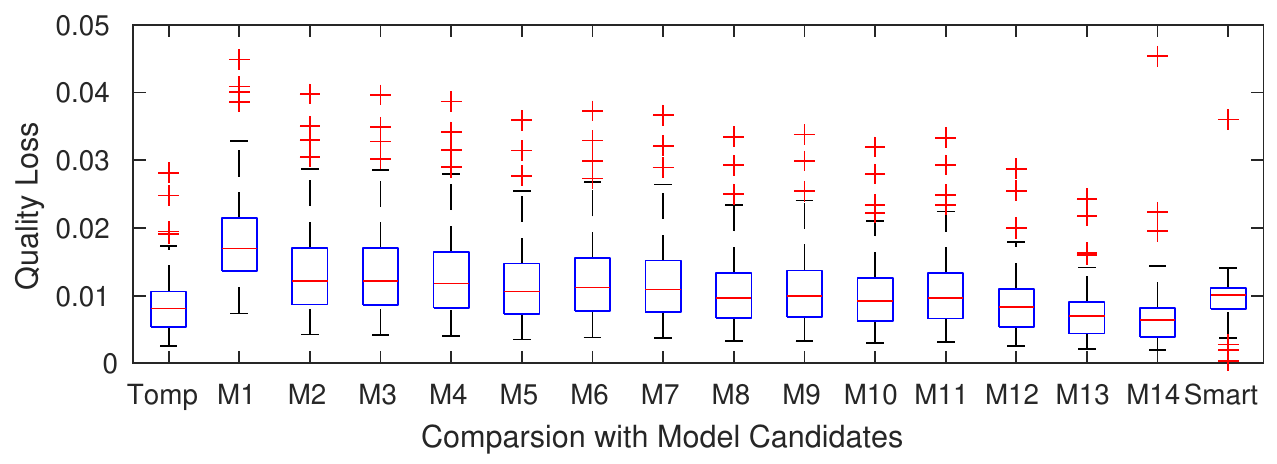}
    \vspace{-10pt}
        \caption{Variation of simulation quality in different model candidates.} 
  \label{fig:sec7_5}
\end{figure}
\noindent\textbf{Speedup}.
We show the performance (the speedup of execution time) of running the 14 neural network models alone without model switching. We use the performance of PCG as the baseline to calculate speedup. 
We also show the performance of Smart-fluidnet and compare it with the 14 individual models. 
Figure~\ref{fig:sec7_4} shows that the performances of the 14 neural network models are quite different, with the speedup ranging from 541.25$\times$ to 141.17$\times$. The performance of Smart-fluidnet is close to the median performance (440.1 $\times$) of the 14 neural network models. This is the result of dynamically using different neural network models at runtime.

\noindent\textbf{Quality}.
We compare the 14 neural network models, the Tompson's model, and Smart-fluidnet, in terms of quality loss. We calculate the quality loss using the method in Section~\ref{sec:model_speedup_accuracy}. Figure~\ref{fig:sec7_5} shows the results. Similar to Figure~\ref{fig:sec7_2} in Section~\ref{sec:model_speedup_accuracy}, the figure shows the distribution and variation using the boxplots. 
Figure~\ref{fig:sec7_5} reveals that the variation of the quality loss in Smart-fluidnet with various input problems is much smaller than any of the 14 neural network models alone. 
With Smart-fluidnet, 91.05\% of the input problems' simulation quality meet the user requirement. With the shortest and longest models (among the 14 neural network models),  12.52\%  and 92.71\% of the input problems' simulation quality meet the user requirement. 

Figures~\ref{fig:sec7_4} and~\ref{fig:sec7_5} include the results for using only the fastest model $M1$ or the most accurate model $M14$ throughout the simulation. $M1$ is $1.18\times$ faster than Smart-fluidnet, but achieves the user requirement on the simulation quality in only 12.52\% of the input problems (for Smart-fluidnet, it is $91.05$);  $M14$ achieves the user requirement on the simulation qualtiy in $92.71\%$ of the input problems, which is close to Smart-fluidnet, but the performance of $M14$ is $3.12\times$ worse than Smart-fluidnet.

Table~\ref{tab:model_distribution} shows the time distribution of five neural network models used by Smart-fluidnet for all input problems. The second row of the table shows the probability of reaching the target when using each neural network model alone, which is predicted by MLP; The third row shows the percentage of execution time of the fluid simulation for the five neural network models (i.e., the time distribution). The table shows that the model with the highest probability, $M7$, takes 50.56\% of the total execution time, which is the longest execution time among the five models.  We also notice that $M5$, which is the fastest model among the five neural network models, takes 18.1\% of the total execution time (the second longest execution time among the five models). These indicate that Smart-fluidnet makes best efforts to reach the user requirement on the simulation quality \textit{and} execution time.

\begin{figure}[!t]
    \centering
    \includegraphics[width=\linewidth, ]{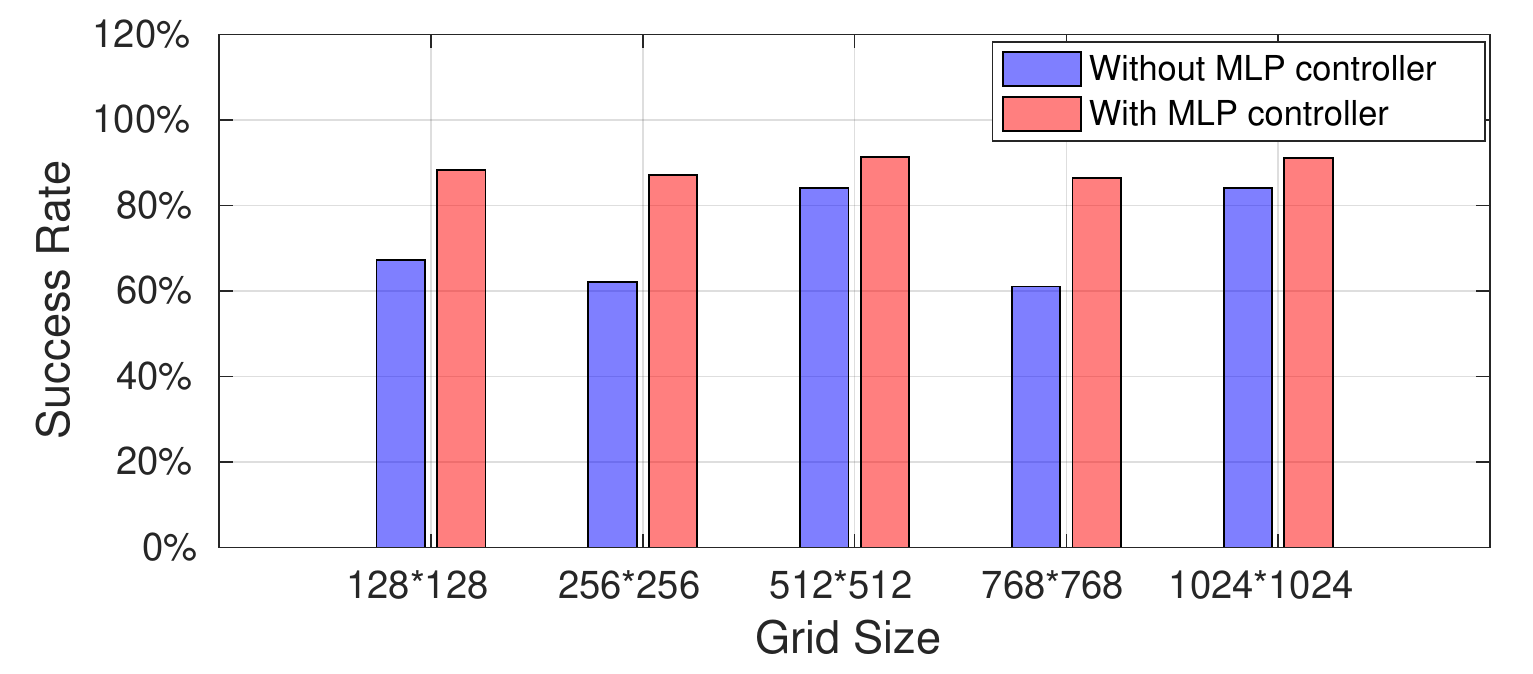}
        \vspace{-25pt}
        \caption{Success rate of reaching target quality with or without using MLP.
        } 
         \vspace{-10pt}
  \label{fig:sec7_6}
\end{figure}

\begin{table}[!t]
\centering

\caption{Execution time distribution for the five neural network models used by Smart-fluidnet at runtime.
}
\vspace{-10pt}
\label{tab:model_distribution}
\begin{tabular}{l p{0.8cm}p{0.8cm}p{0.8cm}p{0.8cm}p{1.2cm}}
\hline
Grid size  & $M7$ & $M5$ & $M10$ & $M2$ & $M13$ \\
\hline   
Prob.(MLP) & 86.12\% & 82.16\% &79.43\% &74.60\% & 70.38\%\\
Time Distr. & 50.56\% & 18.10\% & 11.12\%& 4.07\%& 16.15\% \\
\hline
\end{tabular}
\end{table}

\subsection{Evaluation of MLP Effectiveness}
In this section, we evaluate the effectiveness of MLP. We compare the \textit{success rate} of our runtime system with and without MLP. The success rate means using all input problems for tests (20,480), how many of them reach the simulation quality requirement with our runtime system. Similar to Section~\ref{sec:model_speedup_accuracy}, we use the average quality loss when using all input problems for the fluid simulation with the Tompson's model, as the user requirement.
Without MLP, we have 14 neural network models to be used by the runtime before the fluid simulation, while with MLP, we have five.

Without MLP, we use the fastest neural network model (but less accurate) in the beginning and then switch to more accurate models until we find a model that can reach the user requirement on the simulation quality. We use this model in the remaining of the fluid simulation. Figure~\ref{fig:sec7_6} shows the results. The figure reveals that Smart-fluidnet with MLP causes higher success rates than without MLP. The success rate with MLP is 88.86\% on average and can be up to 91.36\%. This result shows that without MLP, the runtime system can use those neural network models that have a lower possibility to reach the quality target, in order to have better performance. With MLP, we avoid applying those models, hence improving the success rate. We also compare performance with and without MLP. For the grid sizes of 128$*$128, 256$*$256, 512$*$512, 768$*$768, and 1024$*$1024, the corresponding performance when we use MLP, which is normalized by the performance without MLP, is 97\%, 84\%, 92\%, 79\% and 83\% respectively. With MLP, we perform better in all cases.

\subsection{Sensitivity Study: Check Interval}
\label{sec:eva_check_interval}
\begin{figure}[!t]
    \centering
    \includegraphics[width=\linewidth, ]{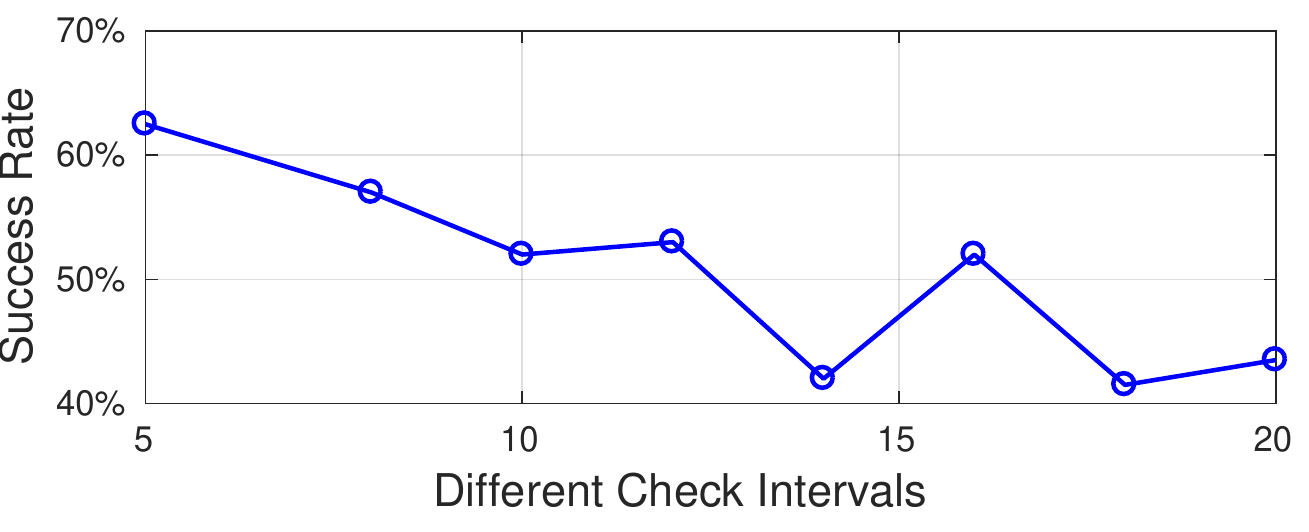}
    \vspace{-15pt}
        \caption{Impact of the check interval on the success rate.
        } 
  \label{fig:sec7_7}
\end{figure}

We study the impact of the check interval on execution time and success rate. We change the check interval, and Figure~\ref{fig:sec7_7} shows the results. Figure~\ref{fig:sec7_7} shows that the success rate decreases when the interval increases. Such decrease is because the model switching is too slow to achieve high simulation quality. We also observe an unusual increase of the success rate when the interval changes from 14 to 16. We attribute such an increase to the statistical variance of using the linear regression method to make the prediction. Although using 16 seems to be useful to increase prediction accuracy for our input problems, using 5 achieves the highest success rate.

Therefore, we use 5 as the check interval throughout our evaluation. We do not use a check interval smaller than 5, because we have to skip the first two time steps to ensure that the growth rate of $CumDivNorm$ is stable and using less than three time steps to build the linear regression model cannot give high prediction accuracy.

\subsection{Evaluation of Resource Usage}

We evaluate resource usage (FLOP and GPU memory consumption) by PCG, the Thompson's and Smart-fluidnet with the grid size of 512*512. Since all input problems for such a grid size has the same resource usage, we randomly choose an input problem and present the evaluation results in Table~\ref{tab:workload}.  

We find that Smart-fluidnet requires much less FLOP than PCG and the Tompson's, which explains why Smart-fluidnet has better performance.
On the other hand, Smart-fluidnet consumes more memory than PCG and the Tompson's, because Smart-fluidnet uses five neural network models on GPU (but not running them simultaneously). However, the memory consumption of Smart-fluidnet is still smaller than the GPU memory capacity (12 GB). If GPU memory is not sufficient, then we may either use fewer neural network models or run them on CPU.

\begin{table}[!t]
\centering
\caption{Resource usage of different methods.}
\vspace{-10pt}
\label{tab:workload}
\begin{tabular}{l p{2.8cm}p{1.8cm}}
\hline
Methods & FLOP (single step)  &GPU Memory  \\
\hline 
PCG & $\sim$1,250 M  &332 MB \\
Tompson & 243.79 M  & 299 MB \\
Smart-fluidnet &110.97 M  & 1,069 MB\\
\hline
\end{tabular}
\end{table}

\section{Related Work}
\label{sec:related_work}
\noindent\textbf{Applying neural networks to HPC applications.}
Neural networks (especially deep neural networks (DNNs)) have been employed in HPC applications recently. In~\cite{richard2014artificial}, a neural network is used to generate input data for modulating the simulation process. Wigley et al.~\cite{sr15:widley} propose a neural network-based online optimization process for the Bose-Einstein condensates (BEC). In~\cite{carli2015incorporating}, neural networks are used to accelerate thermal–hydraulic modeling and enables faster assessment for the dynamic thermal-hydraulic system. Richard et al.~\cite{richard2014artificial} apply a neural network to ITER magnets to predict the occurrence of the interruption, which can be used to adjust the reaction to continue generating power and avoid ITER damage.  Mathuriya et al.~\cite{mathuriya2018cosmoflow} build a CNN model to determine the physical model that describes our universe.

In our work, we use neural network models to approximate computation in an HPC application. Different from the existing efforts, we aim to address the limitation of model flexibility and generality in the existing work~\cite{tompson2017accelerating}.

\noindent\textbf{Acceleration of the fluid simulation.} 
Since the fluid simulation is an important HPC application, many research efforts have been focusing on improving its performance. Popovic et al. use a multi-grid approach to pre-process data for the PCG method in the traditional fluid simulation~\cite{SCA:SCA10:065-073}. 

Molemaker et al. use iterated orthogonal projections and Michael Lentine et al. apply a coarse-grid correction method~\cite{Lentine:2010:NAI:1833349.1778851} to approximate the Poisson's equation~\cite{inproceedings}. These two approaches are effective, but both of them are inexact and only competitive in low-resolution settings. Some recent efforts~\cite{DeWitt:2012:FSU:2077341.2077351} address the above limitation by using a data-driven approach or leveraging some useful statistical characteristics in the data distribution. Our work is different from them, because we use neural network models (not traditional solvers) to improve performance of the fluid simulation.

Some recent efforts attempt to build a neural network model that makes prediction for stability, collisions, forces and velocities of data objects in images or videos~\cite{pmlr-v48-lerer16}. 
Given pressure data from previous frames, voxel occupancy, and velocity divergence, Yang et al. use a patch-based neural network to predict the next pressure~\cite{hong2004precipitation}. 
Jonathan et al. use a convolutional neural network to predict the pressure value in the Eulerian fluid simulation~\cite{tompson2017accelerating}.  
The existing NN-based approximation approaches lack flexibility and generalization, discussed in Section~\ref{sec:intro}.

\noindent\textbf{Model compression.} During the process of model construction, we use some operations to generate simpler neural network models. The recent work using model compression aims to generate simpler models~\cite{cheng2017survey}. There are multiple techniques for model compression, including  
quantization parameters~\cite{jacob2018quantization,wu2018training}, layer pruning~\cite{yang2017designing,molchanov2016pruning}, binarized networks~\cite{rastegari2016xnor,courbariaux2016binarized}, low rank approximation~\cite{tai2015convolutional}, and knowledge distillation~\cite{sau2016deep,lopes2017data}. 
Different from the existing work that focuses on  resource-constrained execution environment (e.g., mobile devices), we focus on simplifying model without resource constrains and study how to build the models to meet the simulation quality for an HPC application.

\vspace{-5pt}
\section{Conclusions}

\label{sec:conclusion}
Using machine learning (especially neural networks) to approximate computation in HPC applications and improve performance has shown preliminary success recently. However, using this approach faces fundamental limitations due to the lack of model flexibility and generality. In this paper, we focus on a specific HPC application and introduce a systematic approach to address the above limitation. In particular, we introduce a framework (Smart-fluidnet) that automatically uses multiple neural network models at runtime to approximate computation and make best efforts to meet the user requirements on simulation quality and execution time. The framework includes a series of techniques to construct and select neural network models. 
Based on the comprehensive evaluation, we show that Smart-fluidnet is $1.46\times$ and $590\times$ faster than a state-of-the-art neural network model and the original fluid simulation respectively on an NVIDIA Pascal GPU, while providing better simulation quality than the state-of-the-art model.

\textbf{Acknowledgement.}
This work was partially supported by U.S. National Science Foundation (CNS-1617967, CCF-1553645 and CCF-1718194) and Chameleon Cloud. We thank reviewers and our shepherd for their constructive comments.

\bibliographystyle{unsrt}

\bibliography{wdong,li,li2}

\end{document}